\documentclass{article}

 \usepackage[main, final]{neurips_2026}
\usepackage[utf8]{inputenc} % allow utf-8 input
\usepackage[T1]{fontenc}    % use 8-bit T1 fonts
\usepackage{hyperref}       % hyperlinks
\usepackage{url}            % simple URL typesetting
\usepackage{booktabs}       % professional-quality tables
\usepackage{amsfonts}       % blackboard math symbols
\usepackage{nicefrac}       % compact symbols for 1/2, etc.
\usepackage{microtype}      % microtypography
\usepackage{xcolor}         % colors

\usepackage{dependencies}

\title{\ourmethod: Structure-Aware Localized Conformal Prediction on Graphs}

\author{%
Peyman Baghershahi\textsuperscript{1} \quad Fangxin Wang\textsuperscript{1} \quad Debmalya Mandal\textsuperscript{2} \quad Sourav Medya\textsuperscript{1} \\
\textsuperscript{1}Department of Computer Science, University of Illinois Chicago \\
\textsuperscript{2}Department of Computer Science, University of Warwick \\
 \textsuperscript{1}\texttt{\{pbaghe2, fwang51, medya\}@uic.edu; \textsuperscript{2}\texttt{debmalya.mandal@warwick.ac.uk}}
}

\begin{document}

\maketitle

\begin{abstract}
  % Conformal prediction (CP) provides a distribution-free approach to uncertainty quantification (UQ) with finite samples. However, applying CP to graph neural networks (\gnns) remains challenging. The combinatorial nature of graphs makes their encoding non-trivial, leading to insufficiently certain prediction logits and indiscriminative embeddings. Existing methods mostly rely on the embedding space for conformal prediction, which can be unreliable for graphs and often yields inefficient prediction sets. We propose \ourmethod, a structure-aware weighted conformal prediction framework that explicitly incorporates the graph topology and inter-node dependencies into localization and weighting. Based on experimental results that show a strong correlation between node homophily and sample coverage, we propose \ourmethod that utilizes a feature-aware densification followed by Personalized PageRank–based kernel computation to model structural proximity between nodes. Accounting for both structural and feature similarity allows \ourmethod to efficiently guarantee marginal coverage while attaining favorable test conditional coverage across extensive experiments on multiple datasets for both regression and classification.
  Conformal prediction (CP) provides a distribution-free approach to uncertainty quantification with finite-sample guarantees. However, applying CP to graph neural networks (\gnns) remains challenging as the combinatorial nature of graphs often leads to insufficiently certain predictions and indiscriminative embeddings. Existing methods primarily rely on embedding-space proximity for localization, which can be unreliable for graphs and yield uneven coverage across different graph regions. We propose \ourmethod, a proximity-based localized CP framework that incorporates graph topology and inter-node dependencies into localization and weighting. Our approach introduces a feature-aware densification step to mitigate locality bias in sparse graphs, followed by a Personalized PageRank-based kernel computation to model structural proximity. This enables topology-dependent anchor sampling and calibration weighting that captures both local and long-range dependencies. Extensive experiments on regression and classification datasets demonstrate that \ourmethod improves test-conditional coverage without excessive conservativeness, while preserving finite-sample marginal coverage.
\end{abstract}

\section{Introduction}
\label{sec:intro}
%%% generic intro

Graph neural networks (\gnns) are increasingly deployed for drug discovery \cite{feinberg2018potential}, weather prediction \cite{ma2023histgnn}, and fraud detection in financial networks \cite{cheng2025graph}. In these high-stakes applications, relying solely on point predictions is risky and the practitioners require reliable uncertainty quantification (UQ) of \gnns to assess the trustworthiness of predictions. Conformal Prediction (CP) is an effective distribution-free framework for constructing set-valued predictions that guarantees covering the ground-truth values using a finite sample of data points \cite{shafer2008tutorial, angelopoulos2021gentle}. Consequently, CP has emerged as a widely adopted approach for post-hoc reliability, applied to graph tasks such as standard regression, link prediction, and out-of-distribution detection \cite{oodcpgnn2025,linkfdr2023, cplinprediction2024}. 

Classical CP guarantees \emph{marginal} coverage on average over the test distribution. However, in many practical settings, uncertainty is heterogeneous, requiring reliability specific to subpopulations or local contexts \citep{validadaptiveacoverage2020, validconfidenseset2021}. Achieving \emph{exact} conditional coverage per sample is known to be impossible in finite samples \cite{infeasiblecondcp2020}. This limitation has motivated a growing body of work on \emph{relaxed} notions of conditional coverage, leveraging localization and weighting strategies to improve validity \citep{wcp2019, beyondexchangeability2022}. A central requirement across these approaches is the availability of a meaningful notion of \emph{proximity} between the test sample and the calibration data.

In graphs, predictive reliability varies across regions with different homophily patterns \citep{naps2023}. Besides, applying CP on graphs is challenging because nodes are naturally dependent, violating standard i.i.d assumptions \citep{networkcp2023, naps2023}. In practice, CP for GNNs often leverages the embedding space, using learned representations as proxies for instances \citep{codrug2023}. However, this proxy can be unreliable as message passing can suffer from a graph-topological bottleneck (over-squashing), restricting global information flow and exacerbating noise under heterophily \citep{oversmoothingoversquashingtradeoff2023}.
Meanwhile, deeper message passing causes over-smoothing, reducing the variance of similarity measures \citep{oversmoothingsurvey2023}. Also, \gnns generally exhibit under-confidence \citep{cagcn2021, gats2022}, suggesting insufficient separability in the learned representation. These effects complicate the core requirement of having a stable proximity kernel for localized CP, which reweights calibration data based on their similarity to the test instance \citep{rlcp2024}. 

Recent work has extended conformal prediction to \gnns to address structural dependencies, with approaches ranging from handling inductive covariate shift \citep{inductivecpgnn2024, naps2023} to developing diffusion-based scores \citep{cpforgnn2023}. Built on findings which preserve exchangeability under permutation-invariant transformation \cite{ranklist2021, networkcp2023, networkassistedcp2025}, recent studies have also applied CP to \gnns in transductive settings \cite{cfgnn2023, rrgnn2025}. Additional related work is provided in the Appendix \ref{app:related_work}.

% However, several limitations remain. Some methods train auxiliary correction or residual models \cite{cfgnn2023, rrgnn2025}, potentially reducing the samples available for calibration, or additionally train the base predictor \cite{codrug2023}. Moreover, they do not fully exploit informative, fine-grained
% topological proximities to capture local variations in prediction uncertainty. Also, the methods based on localized CP typically rely either on \emph{embedding-space} weighting (e.g., kernel density estimation) \cite{codrug2023} or on \emph{purely structural} partitioning (e.g., hard clustering of communities) \cite{rrgnn2025}. These strategies can be fragile because embedding-based similarities may become indiscriminative due to smoothing, while rigid structural clusters may fail to capture fine-grained local heterophily in graphs. As a result, existing graph CP methods typically perform best under strong homophily and may yield overly conservative or poorly \textit{localized prediction sets} in complex topologies. 

However, several limitations remain. Some methods train auxiliary correction or residual models \cite{cfgnn2023, rrgnn2025}, potentially reducing the samples available for calibration, or additionally train the base predictor \cite{codrug2023}. Moreover, these approaches do not fully exploit fine-grained topological proximities to capture local variations in prediction uncertainty. Existing localized methods typically rely on either \emph{embedding-space} weighting \cite{codrug2023} or \emph{purely structural} partitioning \cite{rrgnn2025}. Embedding similarities may become indiscriminative due to smoothing, while rigid structural clusters may overlook local heterophily, yielding overly conservative or poorly localized prediction sets.

% In this work, we bridge the gap between structural and feature-based localization. We propose \ourmethod, a proximity-based localized CP framework for \gnns that explicitly incorporates graph topology and inter-node dependencies into localization-based weighting for conformal quantile computation. Our approach replaces embedding-only proximity with a structure-aware proximity kernel based on Personalized PageRank (PPR), enabling topology-dependent anchor sampling and calibration weighting. To mitigate locality bias in sparse graphs and capture long-range dependencies, we introduce a feature-aware densification step that augments the graph using embedding similarity prior to kernel computation. We further provide theoretical results characterizing how over-smoothing weakens embedding-space separability and how graph connectivity influences structural localization boundaries. 

In this work, we bridge the gap between structural and feature-based localization. We propose \ourmethod, a localized CP framework for \gnns that incorporates graph topology and inter-node dependencies into conformal calibration. Our approach replaces embedding-only proximity with a Personalized PageRank (PPR) kernel for topology-dependent anchor sampling and calibration weighting. To mitigate locality bias in sparse graphs, we introduce feature-aware densification using embedding similarity before kernel computation. We further provide theoretical results characterizing how over-smoothing weakens embedding-space separability and how graph connectivity influences structural localization boundaries. Together, these enable the preservation of marginal coverage while improving conditional coverage in a favorable balance with prediction size. Experiments on seven regression and six classification datasets demonstrate that \ourmethod achieves the target marginal coverage while improving worst-slab coverage without excessive conservativeness.
% The code is available here for reproducibility: \url{https://anonymous.4open.science/r/GraphLCP-D208}.

\section{Preliminaries and Problem Setup}

\subsection{Graph Neural Network (\gnn)} \label{sec:gnns}
In this work, we focus on the canonical form of GNNs, which use a message-passing mechanism. We denote the input graph as $G=(V,E)$, where $V$ and $E$ are its node and edge sets, with $N=|V|$; $\mathbf{X}\in \realset^{N\times d}$ and $\mathbf{A} \in \{0, 1\}^{N\times N}$ are its feature and adjacency matrices. According to \cite{mpnn2017}, an $L$-layer pretrained message-passing model $\phi$ derives a node's $l$-th layer state or representation, $\xv_u^{(l)} (\forall u\in V, \, l \leq L$) using three functions --
(i) a \emph{message-passing} function $M_{\phi}$ that propagates the $l$-th layer state of every node in the neighborhood $\nb_u = \{v: v\in V, \, \mathbf{A}_{uv}=1\}$, (ii) an \emph{aggregate function} $C_{\phi}$ to combine the messages from $M_{\phi}$, and (iii) an \emph{update function} $U_{\phi}$ to derive the node's latest state integrating its $(l-1)$-th layer state and the aggregated message. Formally, $\xv_u^{(l)} = U_{\phi}(C_{\phi}(\{M_{\phi}(\xv_v^{(l)}); v\in \nb_u\}), \xv_u^{(l-1)}).$
Given a downstream task, most GNNs use a simple function (e.g., a linear neural network) to decode the latest embeddings or states to prediction logits as $\hat{y}_u = f(\xv_u^{(L)})$.  Also, for the ease of notation, we use: $\xv_i: \, \forall i \in [N]$ instead of $\xv_u^{(L)}: \, \forall u\in V$, and also $y_i$ instead of $\hat{y}_u$ for $i \in [N]$.

In contrast to standard conformal prediction, where samples are assumed to be exchangeable, conformal validity on graphs is not trivial because node samples are generally dependent, particularly in inductive settings. However, in the node-level transductive setting with a random train/calibration/test split, calibration and test nodes remain exchangeable as long as their labels are not used during training. Once the GNN is trained and frozen, it induces a deterministic transformation from each node and its topological context to an embedding representation, which is then used for conformal prediction. Since conventional GNNs are permutation equivariant and the resulting nonconformity scores are permutation-invariant with respect to calibration and test node permutations, standard conformal prediction guarantees marginal node-level coverage under this transductive setup \cite{cfgnn2023, networkcp2023}.

\subsection{Conformal Prediction (CP)}
\label{sec:CP}

\textbf{Split Conformal Prediction (SCP). }Given a dataset $\data=\{(\xv_i, y_i):(\xv_i, y_i) \sim \dist=\dist_X \times \dist_{Y|X}, \ \xv_i\in \xs, \ y_i\in \ys \}_{i=1}^{N}$ (e.g., $\xv_i$ are the nodes or their embeddings), we first split the dataset to three disjoint sets: 1) $\data_{tr}$ with labels for supervised training, 2) $\data_{cal}$ with labels as holdout calibration set, and 3) $\data_{test}$ purposefully made without labels as test point. After training the message-passing ($\phi$) and decoder ($f$) modules on $\data_{tr}$, we freeze the model for CP. Having $n$ labeled samples $\{(\xv_i, y_i) \in \data_{cal}: i\in [n]\}$, the problem is to construct a prediction interval $\widehat{C}_n(\xv_{n+1})$ for an unlabeled sample $\xv_{n+1} \in \data_{test}$ that contains or ``covers'' the groundtruth label $y_{n+1}$ of the test sample with probability at least $1-\alpha$, where $\alpha$ is the miscoverage rate. 

In order to achieve the desired coverage, every calibration sample and the test sample are assigned a non-conformity as $S_i = s(\xv_i, y_i), \forall i\in [n]$ and $S_{n+1}=s(\xv_{n+1}, y): y\in \ys$, where $s(x, y)$ is a non-conformity score function fixed independently of the calibration and test observations. For example, absolute residual, $s(x, y)=|y-f(x)|$, is a commonly used function. Additionally, let $\hat{q} = \text{Quantile}_{1-\alpha}(\sum_{i=1}^n\frac{1}{n+1}\delta_{s(\xv_i,y_i)}+\frac{1}{n+1}\delta_{+\infty})$, where $\delta_x$ is the probability distribution that places probability $1$ on the value $x$, and $0$ everywhere else. If $S_{n+1}$ is exchangeable with $\{S_i\}_{i=1}^{n}$, then by constructing $\widehat{C}_n(\xv_{n+1})=\{y \in \ys: s(\xv_{n+1}, y) \leq \hat{q}\}$ we have $\prob\{y_{n+1} \in \widehat{C}_n(\xv_{n+1})\} \geq 1-\alpha$ \cite{maincp2005}. This, being averaged over all random draws of the calibration and the test samples, is called \textit{marginal coverage}. However, if the coverage is guaranteed for every specific instance of $\mathbf{x}_{n+1}$, it is called \emph{conditional coverage}, requiring $\mathbb{P}\{y_{n+1} \in \widehat{C}_n(\mathbf{x}_{n+1}) \mid \mathbf{x}_{n+1}\} \geq 1-\alpha$. \cite{infeasiblecondcp2020} shows that achieving such test conditional coverage is infeasible. Therefore, test conditional coverage is commonly relaxed and measured for a test sample $\xv_{n+1} \in B$ where $B \in \mathcal{B}$ and $\mathcal{B}$ is a collection of subsets of feature space, i.e. $\mathcal{B}\subseteq \xs$, and the guarantee requires $\prob\{y_{n+1} \in \widehat{C}_n(\xv_{n+1}) \mid \xv_{n+1} \in B: B\in \mathcal{B}\} \geq 1-\alpha$.

\textbf{Weighted Conformal Prediction (WCP). }Weighted Conformal Prediction (WCP) is one of the ways to attain the above notion of relaxed test conditional coverage \cite{wcp2019}. Suppose we are given a kernel function $H(\xv, \xv')$, e.g. $H(\xv, \xv') = \frac{1}{(2\pi h^2)^{{d}/{2}}}\exp\left(-\frac{\|\xv-\xv'\|^2_2}{2h^2}\right)$. Then we can apply WCP for local coverage by using the kernel centered at $\xv=\xv_{n+1}$, and obtain coverage for a nearby (synthetic) sample $\tilde{\xv}_{n+1}=\xv_{n+1}+h\cdot\mathbf{w}$, where $\mathbf{w}$ is chosen so that $\tilde{\xv}_{n+1}$ follows the feature distribution $\dist_X$ reweighted by $H(\cdot,\xv_{n+1})$. This guarantees coverage for a synthetic test pair $(\tilde{\xv}_{n+1},\tilde y_{n+1})$, where $\tilde y_{n+1}\mid\tilde{\xv}_{n+1}\sim\dist_{Y|X}$, i.e. $\prob\{\tilde y_{n+1}\in\widehat{C}_n(\tilde{\xv}_{n+1})\}\geq1-\alpha$, but not necessarily for $\xv_{n+1}$ itself. For small bandwidth $h$, this provides a locally-weighted relaxation of test-conditional coverage.

\textbf{Randomly Localized Conformal Prediction (RLCP).}
Following WCP, Randomly Localized Conformal Prediction (RLCP)
\cite{rlcp2024} guarantees relaxed test conditional coverage for the test sample $\xv_{n+1}$, rather than a point $\tilde{\xv}_{n+1}$ in its proximity.
Let $H:\xs\times\xs\rightarrow\realset_{\geq 0}$ be
a normalized localization kernel satisfying
$\int_{\xs}H(x,x')\,d\nu(x')=1$ for all $x\in\xs$.
Given a test point $\xv_{n+1}$, RLCP samples an
anchor $\tilde{\xv}_{n+1}\mid\xv_{n+1}
\sim H(\xv_{n+1},\cdot)$ and constructs the
prediction set/interval for the original test point as
$
\widehat C_n
(\xv_{n+1},\tilde{\xv}_{n+1})
=
\left\{
y\in\ys:
s(\xv_{n+1},y)
\leq
\hat q_{1-\alpha}
(\xv_{n+1},\tilde{\xv}_{n+1})
\right\},
$
where
\begin{equation}
\begin{aligned}
&\hat q_{1-\alpha}
(\xv_{n+1},\tilde{\xv}_{n+1})=
\operatorname{Quantile}_{1-\alpha}
\left(
\sum_{i=1}^{n}
\tilde w_i\delta_{s(\xv_i,y_i)}
+
\tilde w_{n+1}\delta_{+\infty}
\right),
\end{aligned}
\label{eq:rlcpquantile}
\end{equation}
with
$
\tilde w_i=
\frac{H(\xv_i,\tilde{\xv}_{n+1})}
{\sum_{j=1}^{n+1}H(\xv_j,\tilde{\xv}_{n+1})}
$,
for $i\in[n+1]$.

Under i.i.d. calibration and test observations, for a pair $(X,\widetilde X)$ s.t. $X\sim\dist_X$ and
$\widetilde X\mid X\sim H(X,\cdot)$, RLCP's relaxed conditional-coverage guarantee
can be characterized by the following theorem.

\begin{thm}[Relaxed conditional coverage of RLCP
\cite{rlcp2024}]
\label{thm:rlcp_conditional}
For any measurable $B\subseteq\xs$ with
$\dist_X(B)>0$, let $B^c=\xs\setminus B$.
Given a metric $d$ on $\xs$, define
$
\operatorname{bd}_{d}(B;r)
:=
\{
x\in B:
\inf_{x'\in B^c}d(x,x')\leq r
\}.
$
Then
\begin{equation}
\begin{aligned}
&\prob\left\{
y_{n+1}\in
\widehat C_n
(\xv_{n+1},\tilde{\xv}_{n+1})
\mid \xv_{n+1}\in B
\right\}\geq 1-\alpha-
\inf_{\delta>0}
\frac{
\prob\{d(X,\widetilde X)>\delta\}
+
\dist_X(\operatorname{bd}_{d}(B;2\delta))
}{
\dist_X(B)
}.
\end{aligned}
\label{eq:rlcp_conditional_bound}
\end{equation}
\end{thm}

Theorem~\ref{thm:rlcp_conditional} bounds the excess conditional miscoverage through: 1) normalized $\prob\{d(\mathbf X,\widetilde{\mathbf X})>\delta\}$, which measures how often the sampled anchor lies outside of its corresponding test point's local neighborhood, and 2) normalized $\dist_X(\operatorname{bd}_{d}(B;2\delta))$, which measures the probability mass of points in $B$ that lie close to its complement $B^c$. Thus, the bound is informative when the localization kernel keeps anchors nearby and the chosen distance preserves separation between $B$ and $B^c$. Motivated by this decomposition, we first study how over smoothing affects the boundary term in GNN embedding space and then examine how this term can be controlled using graph topology.

\subsection{Embedding-space boundary under over-smoothing}
\label{sec:embedding_boundary}

We first investigate the boundary term in
Theorem~\ref{thm:rlcp_conditional} when localization is
defined by distances between GNN embeddings. In
particular, we examine whether over-smoothing causes
nodes in a fixed nonempty proper subset $B\subsetneq V$ to become indistinguishable
from nodes in its complement $B^c$. For the boundary analysis, assume that $G$ is finite, connected,
simple, undirected, and unweighted.
Write $Z=(z_u)_{u\in V}\in\mathbb{R}^{N\times p}$ for its node
representations, where $z_u:=\xv_u$ denotes the frozen GNN embedding
of node $u$. Let $d_Z(u,v):=\|z_u-z_v\|_2$ denote the embedding-induced distance.
We write $\operatorname{bd}_{Z}(B;r):=\operatorname{bd}_{d_Z}(B;r)$, and let $\rho_B=|B|/N$ and $\sigma_1>0$ be the smallest
nonzero eigenvalue of the Laplacian of $G$, and 
$
\mathcal E_G(Z)
:=
\sum_{\{u,v\}\in E}\|z_u-z_v\|_2^2
$ denote the Dirichlet energy of the embedding signal $Z$ on $G$.

\begin{thm}[Embedding-space boundary saturation]
\label{thm:embedding_boundary}
For every embedding matrix $Z$ and every $\delta>0$,
\begin{equation}
\frac{|\operatorname{bd}_{Z}(B;2\delta)|}{|B|}
\geq
\left[
1-
\frac{\mathcal E_G(Z)}
{4\delta^2\rho_B(1-\rho_B)N\sigma_1}
\right]_+,
\qquad [a]_+:=\max\{a,0\}.
\label{eq:embedding_boundary_bound}
\end{equation}
Moreover, let $Z_K\in\mathbb R^{N\times p_K}$ be the node
representations produced by a $K$-layer GNN on the fixed graph
$G$. If $\mathcal E_G(Z_K)\to0$ as $K\to\infty$, then, for
every fixed $B$, $\delta>0$, and sufficiently large $K$
$
\operatorname{bd}_{Z_K}(B;2\delta)=B.
$
\end{thm}

% This theorem isolates the boundary component appearing in the
% conditional-coverage analysis of RLCP~\cite{rlcp2024}. Under the stated
% energy-decay condition, every node in $B$ eventually lies within the
% fixed embedding radius of $B^c$, so embedding distance no longer
% separates these sets at that scale. This is a geometric statement;
% it does not itself establish a conditional-coverage guarantee or a
% finite-depth collapse of the conformal weights. The proof and a
% finite-depth specialization are in Appendix~\ref{app:embedding_boundary}.

% Theorem~\ref{thm:embedding_boundary} shows that, under
% uniform node sampling, over-smoothing (vanishing Dirichlet energy) causes the boundary
% penalty in RLCP's conditional-coverage bound to reach its
% maximum value at any fixed positive embedding radius,
% making the bound uninformative due to the loss of
% geometric separation. 
% A sufficient condition for this energy decay is uniformly
% contractive GNN layers that preserve zero, as established
% in Corollary~\ref{cor:embedding_contractivity} in
% Appendix~\ref{app:embedding_boundary}.
% At finite depth, the theorem also quantifies partial
% boundary saturation when the energy remains positive,
% with stronger guarantees as energy decreases.
% This motivates using graph topology as an
% alternative notion of proximity for localized conformal
% calibration. The proof and a finite-depth specialization
% are provided in Appendix~\ref{app:embedding_boundary}.

Theorem~\ref{thm:embedding_boundary} shows that, under
uniform node sampling, over-smoothing (vanishing Dirichlet
energy) causes the boundary penalty in RLCP's
conditional-coverage bound to reach its maximum value
at any fixed positive embedding radius, making the
corresponding bound uninformative due to the loss of
geometric separation.
As established in Corollary~\ref{cor:embedding_contractivity}
in Appendix~\ref{app:embedding_boundary}, a sufficient condition for this energy decay is that
the GNN layers are uniformly contractive and preserve zero, and at finite depth, the theorem also provides a lower bound
on the boundary fraction when the energy remains positive,
with tighter guarantees as energy decreases.
These findings motivate using graph topology as an
alternative notion of proximity for localized conformal
calibration. The proof of the theorem and further discussion
are provided in Appendix~\ref{app:embedding_boundary}.

\paragraph{Our Objective.}We address the problem of CP for GNNs by explicitly incorporating structural proximity into the weighting step. By decoupling weights from potentially lossy encoders, we create a lower-noise residual pathway that leverages embedding proximity while controlling its influence. \textit{Our goal is to satisfy marginal coverage while achieving strong test-conditional coverage with reasonable prediction sizes.} Theorem~\ref{thm:embedding_boundary} identifies a precise loss of embedding-space separation under energy decay. Our empirical analysis (Appendix~\ref{app:limitations}) examines whether embedding-based localization yields poorly concentrated or nearly uniform weights at the depths and bandwidths used in practice. We use graph topology to supply a complementary notion of proximity.

\section{Our Method: \ourmethod}\label{sec:methods}
\begin{figure}[htbp]
\vspace{-1mm}
    \centering
    \includegraphics[trim={1cm 5cm 0.5cm 0.0cm}, clip, width=0.8\textwidth]{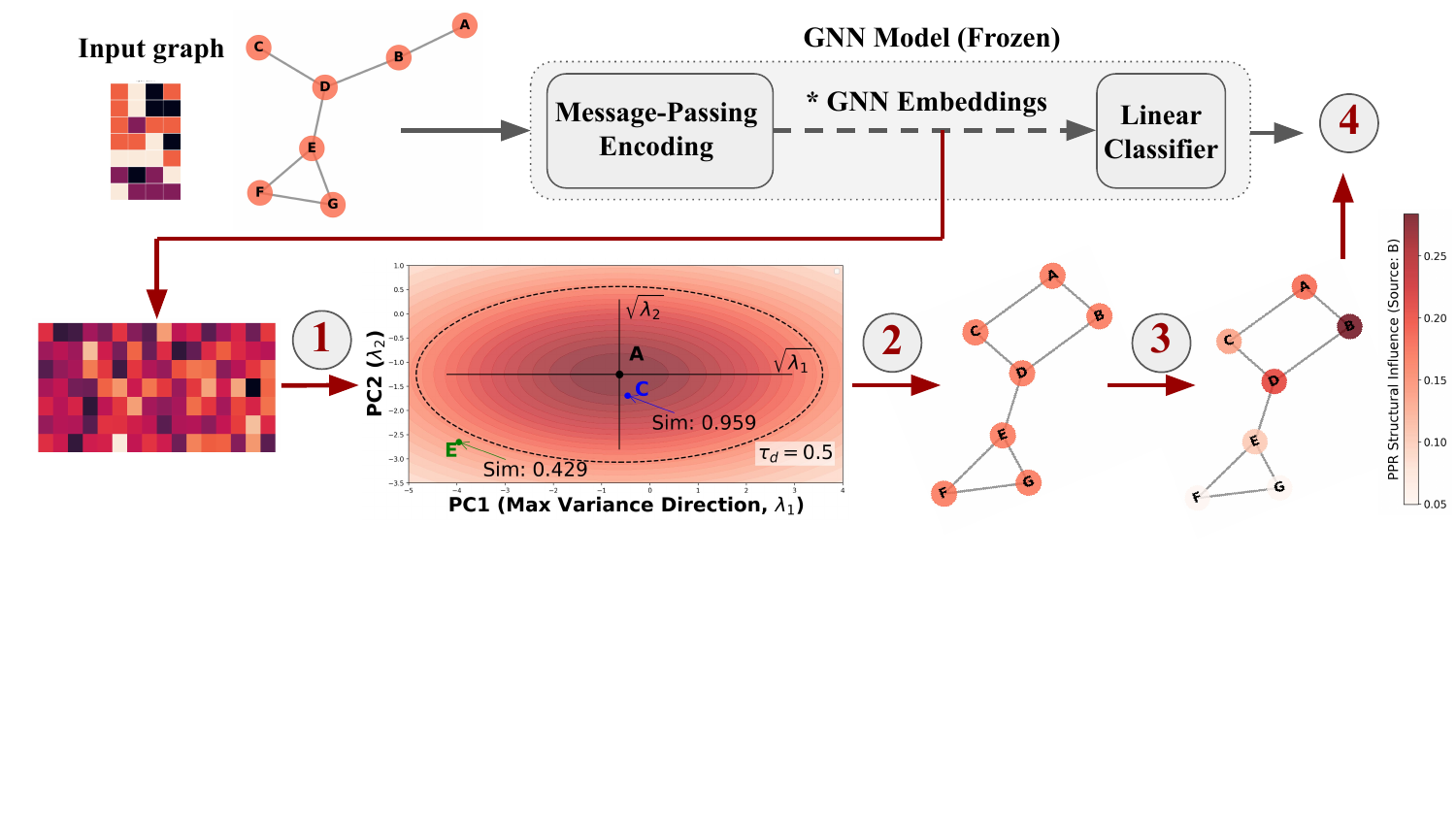}
    \caption{\textbf{\ourmethod}. Given the embeddings of a frozen GNN for an input graph, \ourmethod goes through four stages: 1) \textbf{PCA Transformation}: high-dimensional embeddings are decorrelated and largest eigenvalues (corresponding to directions of highest variance) yield an adaptive bandwidth for an anisotropic Gaussian kernel; 2) \textbf{Graph Densification}: adding informative edges using a fixed similarity threshold; 3) \textbf{PPR Integration}: utilizing the PPR kernel to sample anchor nodes (for each test sample) and weight calibration samples; and 4) \textbf{Conformal Prediction}: a weighted quantile is computed from calibration samples to construct the final prediction set/interval.}
    \label{fig:workflow}
    \vspace{-4mm}
\end{figure}

Recent studies attempted to incorporate structural covariates into GNN-based conformal prediction (CP) \cite{naps2023, rrgnn2025}, yet they have failed to significantly improve conditional coverage. While graph structures are implicitly encoded in \gnn embeddings, lossy encoders or sparse topologies can create bottlenecks, and over-smoothing in denser graphs often diminishes topological dependencies \cite{oversquashingeffect2024, skipconnection2023, oversmoothingsurvey2023}. To mitigate this, we propose \ourmethod, a structure-aware CP method that ensures marginal coverage while achieving high relaxed test conditional coverage. Our approach utilizes a localized kernel and a weighting strategy, illustrated in Figure \ref{fig:workflow}, which remains resilient to embedding suboptimality by explicitly incorporating underlying structural dependencies (empirical evidence demonstrating the strong correlation between node homophily and coverage is provided in Appendix \ref{app:homophily}).

\subsection{Sampling and Weighting with Personalized PageRank (PPR)}\label{sec:ppr_weighting_sampling}
To model structural proximity, we adopt Personalized PageRank (PPR) \cite{ppr2003} as our localization kernel. PPR captures multi-hop influence while maintaining locality via a restart probability, effectively modeling uncertainty propagation in heterogeneous graphs \cite{trustyourneighbor2021}. Unlike shortest-path or k-hop neighborhoods, PPR provides a soft, probabilistic notion of proximity that naturally aligns with weighted CP. Importantly, PPR yields a valid probability distribution over nodes, allowing it to be directly used as a randomized localization kernel compatible with RLCP.

PPR seeded at node $u$ is formally defined as $\ppr_u = \beta \mathbf{e}_u(\mathbf I - (1 - \beta)\mathbf{M})^{-1}$, where $\beta$ is the restart/teleportation probability, $\mathbf{e}_u$ is the one-hot seed vector with $[\mathbf{e}_u]_u=1$ and $[\mathbf{e}_u]_v=0$ for $v\neq u$, and $\mathbf{M}$ is the row-stochastic transition matrix of the weighted graph. PPR yields a vector representing the stationary distribution of a random walk that, at each step, restarts at $u$ with probability $\beta$. The value $[\ppr_u]_v$ denotes the long-run probability of the walk being at node $v$ if started at node $u$. Let $L\sim\operatorname{Geom}(\beta)$ denote a random walk length with $\mathbb P(L=k)=\beta(1-\beta)^k$ for $k=0,1,\ldots$. Then exact PPR is
\begin{equation}
    \ppr_u = \mathop{\expect}_{L \sim \operatorname{Geom}(\beta)} \left[ \mathbf{e}_u \mathbf{M}^L \right] = \sum_{k=0}^{\infty} \beta (1-\beta)^k\mathbf{e}_u \mathbf{M}^k.
\end{equation}

We approximate exact PPR using $K$ restart iterations \cite{appnp2019}, initialized at a seed node $u$ with $\ppr_u^{(0)}=\mathbf e_u$ and updated by $\ppr_u^{(t+1)}=(1-\beta)\ppr_u^{(t)}\mathbf M+\beta\mathbf e_u$. The resulting finite kernel is
\begin{equation}\label{eq:finiteppr}
H_K(u,\cdot)=\ppr_u^{(K)}=\beta\sum_{k=0}^{K-1}(1-\beta)^k\mathbf e_u\mathbf M^k+(1-\beta)^K\mathbf e_u\mathbf M^K.
\end{equation}
To sample an anchor, we draw $L$ as above and return the endpoint of an $L_K=\min(L,K)$-step weighted random walk from $u$ using $\mathbf M$. This yields exactly $H_K(u,\cdot)$, as longer walks are capped rather than discarded, preserving total probability mass. We use the same kernel and transition matrix for anchor sampling and calibration weighting, with $K=30$ in our experiments.

%ensuring that sampling and weighting are governed by the same 

\paragraph{Anchor Sampling. } Analogous to RLCP, in graph space, we construct prediction sets by sampling an anchor node in the neighborhood of the test node and computing conformal scores relative to this anchor. In \ourmethod, anchor nodes are sampled from the finite PPR-induced distribution centered at the test node, so the nearby nodes exert greater influence on the quantile estimation. Given a test node $u_{n+1}$, we sample
\begin{equation}
    \tilde{u}_{n+1}\mid u_{n+1}\sim H_K(u_{n+1},\cdot)=\ppr_{n+1}^{(K)}.
\end{equation}

%\textcolor{red}{[Deb: It would be really helpful to add a pseudoocde for computing the weights.]} \sm{we can also add in appendix if we are out of space}

\textbf{Structural Weighting. }Once the anchor node is sampled, calibration and test nodes are weighted according to their structural proximity to the anchor. Unlike Gaussian kernels, PPR is asymmetric, meaning structural influence is direction-dependent. 
% For undirected graphs, we exploit the degree-normalized symmetry property of PPR to compute weights efficiently without explicitly materializing all PPR vectors.
%However, unlike RLCP which uses a Gaussian kernel, the PPR kernel is asymmetric, we exploit symmetry properties of PPR on undirected graphs to efficiently compute normalized weights.
%Having $\tilde{u}_{n+1}$ sampled, we can weight the calibration and the test nodes. According to RLCP, the kernels used for sampling and weighting must be identical to ensure coverage. However, unlike RLCP, which uses a Gaussian kernel, we use a PPR-based kernel that is not symmetric, i.e., 
Essentially, we have $H_K(u_i, u_j) \neq H_K(u_j, u_i)$, and $\tilde{w}_i \propto H_K(u_i, \tilde{u}_{n+1})$ which requires computing $\ppr_i^{(K)}: \forall i \in [n]$. In cases where we have $n$ calibration samples and $m$ test samples s.t. $n<m$, it is efficient to compute $\ppr_i^{(K)}: \forall i \in [n]$ directly. However, when $n>m$ (e.g., in a single-sample/pointwise case), this approach is inefficient. To resolve this issue for undirected graphs, we compute weights efficiently by exploiting the degree-normalized symmetry property, which also holds for the finite kernel: $d_iH_K(u_i, u_j) = d_jH_K(u_j, u_i)$, where $d_i$ denotes the weighted degree of $u_i$ in the graph used for PPR. In particular, after densification and isolated-node handling, $d_i=\tilde{\mathbf D}_{ii}$. Thus $d_i[\ppr_i^{(K)}]_j=d_j[\ppr_j^{(K)}]_i$. Letting $\tilde{\ppr}_{n+1}=\ppr_{\tilde u_{n+1}}^{(K)}$ denote the finite PPR distribution seeded at the sampled anchor $\tilde{u}_{n+1}$, the calibration and test nodes are weighted according to the anchor node as follows %\sm{lot of space around eq}:
\begin{equation}\label{eq:pprweights}
    \tilde{w}_i = \frac{ d_i^{-1} [\tilde{\ppr}_{n+1}]_i }{ \sum_{j=1}^{n+1} d_j^{-1} [\tilde{\ppr}_{n+1}]_j }, \quad \forall i\in [n+1]
\end{equation}
Intuitively, nodes that exert stronger structural influence on the anchor node contribute more to the conformal quantile, adapting the graph context. This topological control of nodes' contributions particularly helps with conditional coverage by filtering out noisy information and focusing on structurally proximal nodes. 

Under the transductive random-split setting, permutation-invariant nonconformity scores preserve the exchangeability of calibration and test nodes \cite{cfgnn2023}. Furthermore, conditioning on the sampled anchor reweights the probability of each node being the test node proportionally to its localization kernel value. Since GraphLCP uses these same kernel values for weighted calibration, the weighted conformal argument yields marginal coverage, provided the kernel and scores are fixed independently of calibration/test roles and labels.

\textbf{Remark.} While the PPR-based kernel effectively captures structural similarity between nodes, it does not incorporate node attributes. At the same time, although GNN embeddings are imperfect and can be noisy due to information bottlenecks and over-smoothing, they determine the prediction logits and therefore remain an important source of information for uncertainty quantification. In addition, the PPR-based proximity measure behaves differently depending on graph density. In dense graphs, PPR explores many paths via random walks and captures global structure, whereas in sparse graphs it performs poorly due to strong localization \cite{stronglocalizationppr2015}, since the random walk probability mass drops rapidly, leading to negligible scores for distant nodes. This causes biased or overly localized weighting. 

To address these limitations, we introduce feature-aware graph densification before computing the PPR kernel. Theorem~\ref{thm:embedding_boundary} shows that, if Dirichlet energy tends to zero, every node in $B$ eventually becomes close in embedding space to $B^c$ at any fixed positive radius. At finite depth, we hypothesize that relative embedding similarities can still identify useful edges even as absolute distances shrink. We therefore use thresholded similarities to augment the topology, while PPR on the augmented graph determines anchor sampling and calibration weights. We also hypothesize that this multi-hop weighting is less sensitive to individual noisy similarities than directly weighting in continuous embedding-space. Our boundary analysis motivates the use of graph-space localization, and we evaluate the weighting strategy empirically.

\subsection{Densification to Address Local Bias}

We densify the graph before applying the kernel through three stages: (i) PCA fitted on pooled unlabeled calibration and test embeddings to concentrate informative signals and reduce feature correlations, (ii) an anisotropic Gaussian similarity kernel with adaptive bandwidths, and (iii) informative edge additions using a fixed similarity threshold. These steps expand the effective neighborhood of the structural kernel while limiting noise and over-densification.

For densification, we exploit the embedding similarities in the encoder embedding space using a Gaussian kernel. This kernel has two advantages: (i) unlike other similarity measures (e.g., cosine), it captures the non-linearity of the embeddings; and ii) the bandwidth controls the range of effect to adjust for local or global densification. To densify the graph, we construct a new adjacency matrix $\tilde{\mathbf{A}}$:
\vspace{-1mm}
\begin{equation}\label{eq:denseadj}
[\tilde{\mathbf{A}}]_{ij} =
\begin{cases} 
1 & \text{if } [\mathbf{A}]_{ij}=1 \\
k_h(\xv_{i}, \xv_{j}) & \text{if } i\neq j,\ [\mathbf{A}]_{ij}=0 \text{ and } k_h(\xv_{i}, \xv_{j}) \geq \tau_d \\
0 & \text{o.w.}
\end{cases}
\end{equation}

%\textcolor{red}{[The definition is confusing. Do you check $[A]_{ij} = 1$ or $k_h(\xv_i, \xv_j) \ge \tau_d$? What if both are true simultaneously?]}\pb{Is is clear now?}

% where $k_h(\xv_i,\xv_j)$ is the anisotropic Gaussian similarity on the PCA-projected embeddings defined below, and $\tau_d$ is a fixed similarity threshold specified before calibration and test evaluation. 
where $k_h(\xv_i,\xv_j)$ is the anisotropic Gaussian similarity
on PCA-projected embeddings defined below, and $\tau_d$ is a
fixed similarity threshold set before calibration and evaluation.
For nodes with zero weighted degree, we add a self-loop before computing the PPR transition matrix. With $\tilde{\mathbf{D}}_{ii} = \sum_{j=1}^{N}[\tilde{\mathbf{A}}]_{ij}$ recomputed after this step, the transition matrix is $\mathbf{M}=\tilde{\mathbf{D}}^{-1}\tilde{\mathbf{A}}$, which is row-stochastic.

% The densification threshold $\tau_d$ effectively determines the number of edges added. Small threshold values result in a nearly complete graph, increasing computational time. Moreover, the message-passing embeddings generated may be insufficiently discriminative; thus, smaller thresholds may allow the addition of potentially noisy edges. Conversely, large thresholds may hinder the addition of new edges, thereby missing informative edges that can overcome the sparsity barrier in applying the PPR-based kernel. We fix $\tau_d$ before calibration without using calibration or test labels. The resulting adjacency is used for both anchor sampling and calibration weighting. PCA uses pooled unlabeled calibration and test embeddings, preserving the calibration/test role symmetry of the feature-based graph construction.

The densification threshold $\tau_d$ controls the number of added
edges. Small thresholds may produce nearly complete graphs,
increasing computational cost and introducing noisy edges when
message-passing embeddings are insufficiently discriminative.
Conversely, large thresholds may miss informative connections
needed to overcome graph sparsity. We fix $\tau_d$, and use
the resulting adjacency for both anchor sampling and calibration
weighting. PCA is fitted on pooled unlabeled calibration and test
embeddings, preserving calibration/test role symmetry. Find more details on the practical design principles of PCA in Appendix~\ref{app:pca_application}.

\paragraph{Dimensionality Reduction.}

Two pivotal issues with \gnn embeddings are as follows: (i) dimensional information redundancy \cite{resisitoversmoothing2024}, i.e., feature dimensions are highly correlated, and (ii) the lack of discriminative power due to over-smoothing \cite{oversmoothingsurvey2023}. These accumulate noise and errors in similarity measures derived from the raw, high-dimensional, ill-posed space. 

We use PCA \cite{pca2014} to decouple feature correlations and condense the embedding space. Let $V_{\mathrm{pool}}=V_{\mathrm{cal}}\cup V_{\mathrm{test}}$ and $N_{\mathrm{pool}}=|V_{\mathrm{pool}}|$. Using their unlabeled embeddings, let $\bar{\xv}_{\mathrm{pool}}=N_{\mathrm{pool}}^{-1}\sum_{i\in V_{\mathrm{pool}}}\xv_i$, and let $\mathbf Z_{\mathrm{pool}}\in\mathbb R^{N_{\mathrm{pool}}\times p}$ contain the centered rows. Set $\mathbf C_Z=N_{\mathrm{pool}}^{-1}\mathbf Z_{\mathrm{pool}}^T\mathbf Z_{\mathrm{pool}}$. We choose $c$ leading orthonormal eigenvectors $\mathbf V_c\in\mathbb R^{p\times c}$ of $\mathbf C_Z$, with eigenvalues $\lambda_1\geq\cdots\geq\lambda_c>0$, and write $\mathbf\Lambda_c=\operatorname{diag}(\lambda_1,\ldots,\lambda_c)$. The projected embedding of any node $i$ is $\mathbf h_i=\mathbf V_c^T(\xv_i-\bar{\xv}_{\mathrm{pool}})$. These embeddings are used to compute similarities for densification. Retaining high-variance directions reduces redundancy but cannot recover distinctions lost to complete over-smoothing.

\paragraph{Selecting Informative Edges by Adaptive Bandwidth.}
Even in the reduced space, samples are not uniformly distributed: topological bottlenecks often make embeddings dense in some directions and sparse in others. Proximity in such non-uniform distributions cannot be effectively captured by an isotropic Gaussian kernel, which assumes equal variance across all dimensions. In contrast, an anisotropic kernel accounts for varying scales across dimensions and captures geometric correlations. Therefore, we construct an anisotropic Gaussian kernel with adaptive bandwidth.

The PCA projection offers a specific advantage: the variance of the projected data along each principal component equals its corresponding eigenvalue. This motivates us to adaptively set the bandwidths of the anisotropic Gaussian kernel based on the PCA eigenvalues. Formally, our Gaussian kernel in the PCA projection space is defined as:
\begin{equation}\label{eq:adaptivebandwidth}
k_h(\xv_i,\xv_j) = \exp\left(-\frac{1}{2}D_M^2(\mathbf h_i,\mathbf h_j;h^2\mathbf\Lambda_c)\right)
\end{equation}
where $D_M^2(a,b;h^2\mathbf\Lambda_c)=(a-b)^T(h^2\mathbf\Lambda_c)^{-1}(a-b)$ is the squared Mahalanobis distance on the projected space and $h>0$ is the base bandwidth. The retained eigenvalues are positive, ensuring a well-defined inverse. Without PCA, Gaussian similarities use the original embeddings.

\textbf{Weighted Quantile Computation and Prediction. } After graph densification, anchor point sampling, we apply the structure-aware weighting as in Eq. \ref{eq:pprweights} and use the weights for computing the weighted quantile $\hat{q}_{1-\alpha}(\xv_{n+1}, \tilde{\xv}_{n+1})$ following Eq.\ref{eq:rlcpquantile}. Finally we derive the prediction set or interval as $\widehat{C}_n^{\text{\ourmethod}}(\xv_{n+1}, \tilde{\xv}_{n+1}) = \{y\in \ys: s(\xv_{n+1}, y) \leq \hat{q}_{1-\alpha}(\xv_{n+1}, \tilde{\xv}_{n+1})\}$. 

% \subsection{Graph-space boundary control}

% The preceding embedding result motivates studying localization in graph
% space. We bound the graph-space boundary of the same fixed node set
% $B$ to identify when topology can retain a smaller boundary at a
% chosen integer radius.
% For vertices $u,v\in V$, let $d_G(u,v)$ denote the
% length of the shortest path from $u$
% to $v$ within $G$. For a nonempty set $S\subseteq V$, define
% $
% d_G(u,S):=\min_{v\in S}d_G(u,v).
% $
% Define the graph-space inner
% boundary, for an integer $r\geq1$, by
% $$
% \operatorname{bd}_G(B;r)
% :=
% \{u\in B:d_G(u,B^c)\leq r\}.
% $$

% Also, the subgraph induced by $B$ is
% $
% G[B]=(B,E[B]),
% $
% where 
% $
% E[B]:=\bigl\{\{u,v\}\in E:u,v\in B\bigr\}.
% $
% Thus, $G[B]$ contains every vertex in $B$ and every edge
% of $G$ whose endpoints both belong to $B$.
% If two vertices are disconnected in $G[B]$, their distance in
% $G[B]$ is defined as $+\infty$.
% For $t\geq0$, let
% $$
% \Gamma_B(t)
% :=
% \max_{\varnothing\neq S\subseteq B}
% \frac{
% |\{u\in B:d_{G[B]}(u,S)\leq t\}|
% }{|S|}.
% $$
% Thus $\Gamma_B(t)$ measures the largest $t$-step
% neighborhood growth within $G[B]$.

\subsection{Graph-space boundary control}
\label{sec:graph_boundary}

The preceding embedding-space result motivates studying
the boundary term in RLCP's conditional-coverage bound
using graph topology instead of GNN embedding distances.
For vertices $u,v\in V$, let $d_G(u,v)$ denote their
shortest-path distance in $G$, and let
$d_G(u,S):=\min_{v\in S}d_G(u,v)$ for nonempty
$S\subseteq V$. Following
Theorem~\ref{thm:rlcp_conditional}, we define
$\operatorname{bd}_G(B;r):=
\operatorname{bd}_{d_G}(B;r)$ for integer $r\geq1$.
Let $G[B]=(B,E[B])$ be the subgraph induced by $B$,
where $E[B]=\{\{u,v\}\in E:u,v\in B\}$.

Distances between disconnected vertices in $G[B]$
are taken as $+\infty$. For $t\geq0$, define
\[
\Gamma_B(t)
:=
\max_{\varnothing\neq S\subseteq B}
\frac{
|\{u\in B:d_{G[B]}(u,S)\leq t\}|
}{|S|},
\]
which measures the maximum $t$-step neighborhood
growth within $G[B]$. Let $E(B,B^c) = \bigl\{\{u,v\}\in E:u\in B,\ v\in B^c\bigr\}$ denote the set of edges crossing
between $B$ and $B^c$.
% and
% $
% \deg_G(u):=|\{v\in V:\{u,v\}\in E\}|.
% $
% Write $\operatorname{vol}_G(S):=\sum_{u\in S}\deg_G(u)$.
% The conductance of $B$ in $G$ is
% $$
% \Phi_G(B):=
% \frac{|E(B,B^c)|}
% {\min\{\operatorname{vol}_G(B),
%        \operatorname{vol}_G(B^c)\}}.
% $$

\begin{thm}[Graph-space boundary control]
\label{thm:graph_boundary}
For every integer $r\geq1$,
\begin{equation}
\begin{aligned}
\frac{|\operatorname{bd}_G(B;r)|}{|B|}
&\leq
\min\left\{
1,\,
\frac{\Gamma_B(r-1)|E(B,B^c)|}{|B|}
\right\}.
% \\
% &=
% \min\left\{
% 1,\,
% \frac{\Gamma_B(r-1)\Phi_G(B)
% \min\{\operatorname{vol}_G(B),
%        \operatorname{vol}_G(B^c)\}}{|B|}
% \right\}.
\end{aligned}
\label{eq:graph_boundary_bound}
\end{equation}
\end{thm}

Unlike Theorem~\ref{thm:embedding_boundary}, which
shows boundary saturation under over-smoothing,
Theorem~\ref{thm:graph_boundary} upper-bounds the
boundary penalty in RLCP's conditional-coverage analysis.
Under uniform node sampling, a small crossing cut and
moderate neighborhood growth keep this penalty small.
The localization-tail penalty is also controlled by PPR:
since an anchor $\tilde u\mid u\sim H_K(u,\cdot)$ is sampled
through a random walk of length
$L_K=\min(L,K)$ with $L\sim\operatorname{Geom}(\beta)$,
$\prob\{d_{\tilde G}(u,\tilde u)>t\mid u\}
\leq(1-\beta)^{t+1}$ for any integer $t\geq0$.
Together, these results bound both penalties in
Eq.~\ref{eq:rlcp_conditional_bound}, motivating
structural localization. But, they do not independently
establish conditional-coverage guarantees in the
transductive setting.

% Moreover,
% Proposition~\ref{prop:ppr_tail} bounds the localization-tail
% term through the PPR restart probability $\beta$.

For \ourmethod, both results are applied to the
unweighted graph of the densified adjacency
$\tilde{\mathbf A}$ in Eq.~\ref{eq:denseadj},
excluding self-loops. Proofs and an explicit maximum-degree bound are in Appendix~\ref{app:graph_boundary_proof}.
Algorithm \ref{alg:ourmethod} in Appendix \ref{app:pseudocode} presents the pseudocode of our proposed method \ourmethod.

\section{Experiments} 
\label{sec:experiments}

\textbf{Datasets. }We show performance across \textit{seven node regression} datasets, and \textit{six node classification} datasets, and provide their details and statistics in the Appendix \ref{app:datasets}.\\
\textbf{Metrics. }We show marginal coverage as a conventional metric. However, as discussed in Sec. \ref{sec:methods}, our localization-based approach is more advantageous for efficient relaxed conditional coverage. Therefore, for conditional CP, we use two settings from the recent comprehensive work on conditional CP diagnostics \cite{conditionalcp2025}. 1) Worst-case diagnostics: we evaluate worst-slab coverage (WSC) \cite{validadaptiveacoverage2020, validconfidenseset2021}. 2) Group-based diagnostics: we show \textit{four} feature-based groupings (partitioning).\\
\textbf{Baselines. }For validity, we compare with various baselines: vanilla split CP (SCP), two localized CP methods, namely RLCP \cite{rlcp2024} and CalLCP \cite{callcp2022}, and five graph-specific methods, specifically NAPS \cite{naps2023}, DAPS \cite{cpforgnn2023}, CF-GNN\cite{cfgnn2023}, and SNAPS \cite{snaps2024}, and RR-Mondrian \cite{rrgnn2025}. For a fair comparison, we use the standard nonconformity scoring mechanism of vanilla split CP across methods, replacing the APS-based mechanism \cite{validadaptiveacoverage2020} in NAPS, DAPS, SNAPS, and CF-GNN. Alternative approaches such as APS for classification and CQR \cite{cqr2019} for regression can independently affect coverage and prediction size, making it difficult to isolate the contributions of the underlying calibration mechanisms. CalLCP’s high computational cost on large graphs exceeded our 3-day time limit, so it is occasionally excluded from our results.  We fix the miscoverage rate to $\alpha = 0.1$ across all experiments. Additional details on other settings are provided in Appendix \ref{app:experiments}. Code is available at \href{https://github.com/pbaghershahi/GraphLCP.git}{https://github.com/pbaghershahi/GraphLCP.git}.

\begin{figure}[t]
    \centering
    % width=0.8\textwidth (or whatever size you prefer)
    \includegraphics[width=0.95\linewidth]{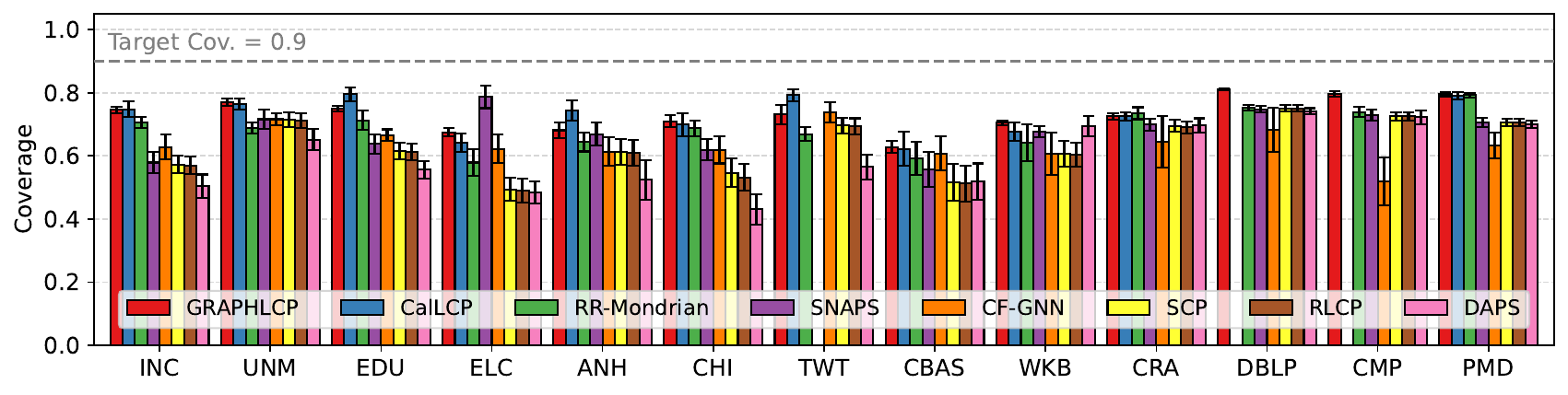}
    \caption{Results for WSC on regression and classification datasets. Miscoverage rate $\alpha=0.1$. Overall, \ourmethod improves WSC over SCP and RLCP across all datasets.}
    \label{fig:barwsc}
    \vspace{-3mm}
\end{figure}

\subsection{Results on Conditional and Marginal CP}

\subsubsection{Conditional Coverage}
We evaluate the conditional coverage under worst-case and group-based settings. Note that the worst-case setting using \textit{WSC is the most challenging evaluation}, as it is an optimization-based evaluation that adversarially seeks a bounded interval along a projection direction where a sufficient number of test points exhibit the minimum coverage. Therefore, searching across a large number of directions could reveal the method's lowest conditional coverage.

\textbf{Worst-case Setting. } We take $100$ random direction vectors sampled uniformly from the unit sphere. The results are presented in Figure \ref{fig:barwsc} for both classification and regression datasets. \ourmethod improves WSC over SCP and RLCP across all datasets and remains competitive with the strong baseline CalLCP, outperforming it on six of the eleven datasets with available results (Table \ref{tab:wsc_coverage_all}). It also performs favorably against other graph-specific methods. NAPS attains high WSC on classification datasets but often with much larger sets and marginal overcoverage. These results highlight the benefit of structure-aware localization and feature-aware densification. 

\begin{wrapfigure}[19]{r}{0.5\textwidth} 
\vspace{-0.2cm}
    \centering
    \includegraphics[
        width=0.95\linewidth, % Use width relative to the wrapfigure box
        trim={0.2cm 0.2cm 0.2cm 0.2cm}, 
        clip
    ]{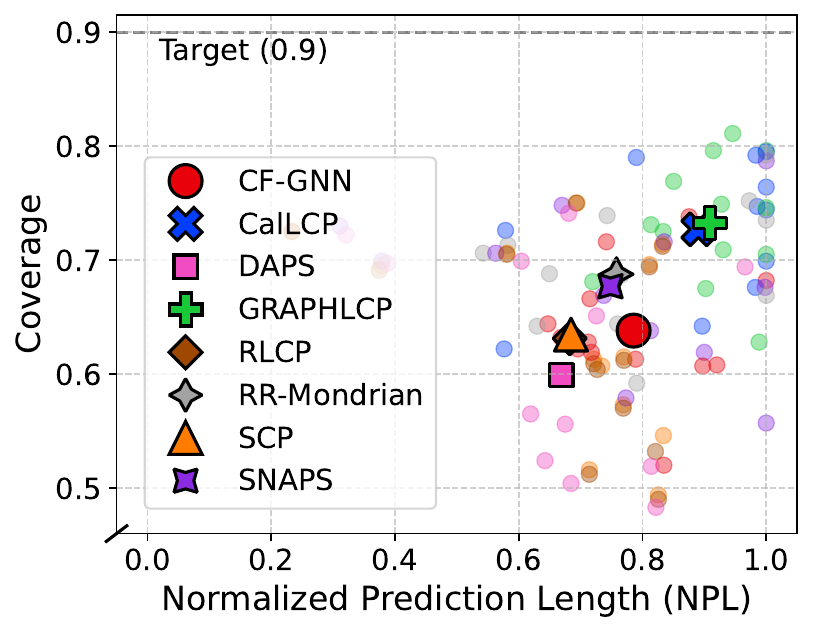}
    \caption{Results for WSC and efficiency on all datasets. \ourmethod achieves the highest average WSC among the methods shown, with slightly larger NPL than CalLCP.}
    \label{fig:scatterwsc}
\end{wrapfigure}

Next, we compare WSC and prediction length across datasets in Figure \ref{fig:scatterwsc}, along with each method's centroid of performance. Prediction lengths (sizes) are normalized by the maximum length across methods (denoted by NPL). \ourmethod achieves the highest average WSC, with slightly larger normalized prediction lengths than CalLCP. This demonstrates the improvement in local coverage
despite a modest efficiency cost. An additional
analysis in Appendix~\ref{app:localization_tradeoff}
shows that increasing structural localization
consistently improves WSC with a gradual increase
in prediction size.

%\begin{wrapfigure}[17]{r}{0.5\textwidth} 
\begin{wrapfigure}[16]{r}{0.5\textwidth} 
% The [12] tells LaTeX to wrap exactly 12 lines of text
% {r} for right side, {l} for left
    \vspace{-0.8cm}
    % \vspace{-0.4cm}
    \centering
    \includegraphics[
        width=0.95\linewidth, 
        trim={0.2cm 0.2cm 0.2cm 0.0cm}, % trim={<left> <bottom> <right> <top>}
        clip                    % 'clip' is required for trim to work
    ]{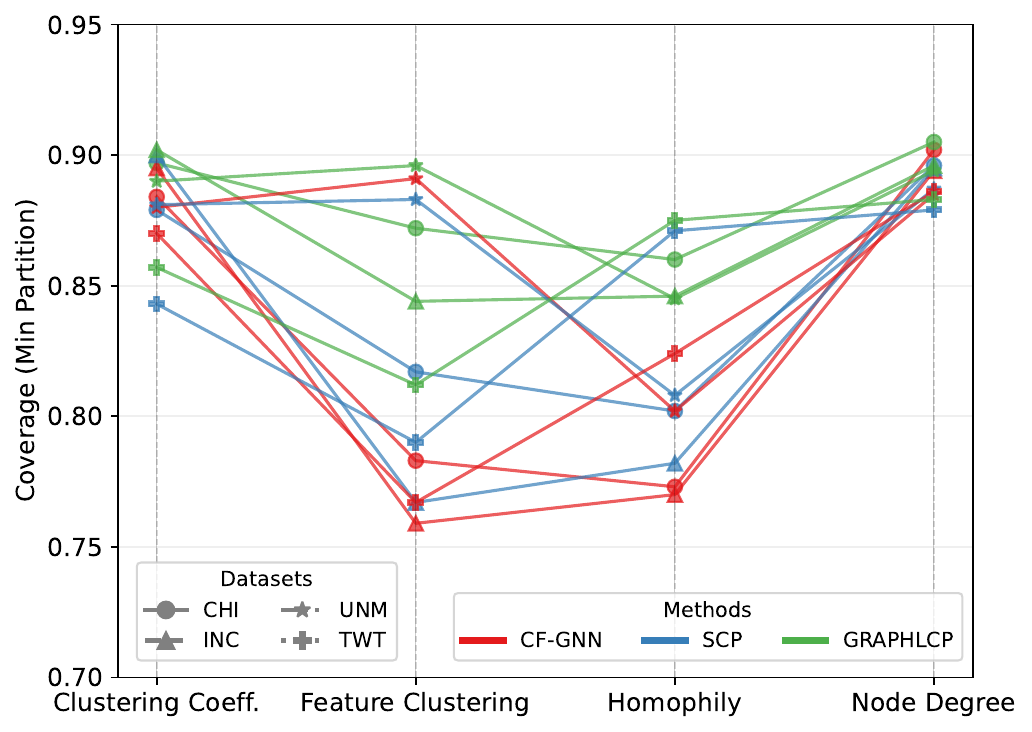}
    
    \caption{Group-based conditional coverage results ($\alpha=0.1$), showing minimum coverage across semantic and structural partitions. \ourmethod outperforms baselines, particularly on homophily and feature-clustering partitions.}
    \label{fig:fscparallel}
    % \vspace{-1cm}
\end{wrapfigure}

% \textbf{Group-based Setting. } Next, we evaluate group-based metrics, where we partition or group the test samples, based on different semantics, as of \textit{Feature Clustering} based on the message-passing embeddings, and structural, as of \textit{Clustering Coefficient}, \textit{Homophily}, and \textit{Node Degree}. This enables us to test the methods on the feature-stratified conditional coverage \cite{conditionalcp2025}. For each feature, we partition the data into three equal quantiles: ``Low'', ``Mid'', and ``High''.  We select four datasets that allow this 3-way partitioning, and bring the worst coverage across all partitions in Figure \ref{fig:fscparallel} for every feature. The general pattern shows the superiority of \ourmethod and is consistent with the results from the WSC experiments. 

\textbf{Group-based Setting. } Next, we evaluate group-based conditional coverage by partitioning test samples based on message-passing embeddings (\textit{Feature Clustering}) and structural properties (\textit{Clustering Coefficient}, \textit{Homophily}, and \textit{Node Degree}) \cite{conditionalcp2025}. For each property, we divide samples into three equal quantiles: ``Low'', ``Mid'', and ``High''. Figure~\ref{fig:fscparallel} reports the worst coverage across partitions on four selected datasets. The results consistently favor \ourmethod, aligning with the WSC findings.

The effect of \ourmethod's two major components is evident in the group-based conditional coverage results, where we consistently outperform. By leveraging structural information, motivated by the importance of homophily in graph-based CP (Appendix \ref{app:homophily}), \ourmethod improves coverage on homophily-based partitions while also outperforming baselines on feature-based partitions.

\subsubsection{Marginal Coverage}
% Evaluating marginal coverage is a sanity check for a CP method. Results in Table \ref{tab:marginal_coverage_all} show that most methods meet or nearly meet the target coverage, although DAPS undercovers on several regression datasets. Since marginal coverage does not reveal local undercoverage, WSC remains important for distinguishing their conditional performance. We present marginal prediction length results in Appendix \ref{app:addiotnalexperiments}.
Marginal coverage serves as a sanity check for CP methods.
Table~\ref{tab:marginal_coverage_all} shows that most methods
meet or approach the target, although DAPS undercovers on
several regression datasets. Since marginal coverage can
mask local undercoverage, WSC remains important for assessing
conditional performance. Marginal prediction lengths are
reported in Appendix~\ref{app:addiotnalexperiments}.

\begin{table*}[t]
\scriptsize
\centering
\caption{Marginal coverage results. The target max miscoverage of $\alpha=0.1$ is almost met by the baselines (except for DAPS). ``-'' means infinite-length prediction interval. OOT (out-of-time) takes an unreasonably long time.}
\vspace{-1mm}
\resizebox{\textwidth}{!}{
    \begin{tabular}{L{0.6in}C{0.3in}C{0.3in}C{0.3in}C{0.3in}C{0.3in}C{0.3in}C{0.3in}C{0.3in}C{0.3in}C{0.3in}C{0.3in}C{0.3in}C{0.3in}}
        \toprule
        \textbf{Method} & \textbf{ANH} & \textbf{CHI} & \textbf{EDU} & \textbf{ELC} & \textbf{INC} & \textbf{UNM} & \textbf{TWT} & \textbf{CMP} & \textbf{CRA} & \textbf{DBLP} & \textbf{CBAS} & \textbf{WKB} & \textbf{PMD} \\
        \midrule
        SCP & 0.90\std{.01} & 0.90\std{.01} & 0.91\std{.01} & 0.90\std{.01} & 0.90\std{.00} & 0.91\std{.01} & 0.90\std{.01} & 0.90\std{.00} & 0.90\std{.00} & 0.90\std{.00} & 0.90\std{.01} & 0.90\std{.02} & 0.90\std{.00} \\
        NAPS & - & - & - & - & - & - & - & 0.93\std{.00} & 0.99\std{.00} & 0.99\std{.00} & 1.00\std{.00} & 1.00\std{.00} & 0.99\std{.00} \\
        DAPS & \textcolor{red}{0.87\std{.02}} & \textcolor{red}{0.86\std{.01}} & \textcolor{red}{0.88\std{.01}} & 0.90\std{.01} & \textcolor{red}{0.87\std{.01}} & \textcolor{red}{0.87\std{.01}} & \textcolor{red}{0.82\std{.01}} & 0.90\std{.00} & 0.90\std{.00} & 0.90\std{.00} & 0.90\std{.01} & 0.90\std{.01} & 0.90\std{.00} \\
        RLCP & 0.90\std{.01} & 0.90\std{.01} & 0.90\std{.01} & 0.90\std{.01} & 0.90\std{.00} & 0.90\std{.01} & 0.90\std{.01} & 0.90\std{.00} & 0.90\std{.00} & 0.90\std{.00} & 0.90\std{.01} & 0.90\std{.02} & 0.90\std{.00} \\
        CF-GNN & 0.92\std{.01} & 0.91\std{.01} & 0.90\std{.01} & 0.90\std{.01} & 0.90\std{.01} & 0.90\std{.01} & 0.91\std{.01} & 0.90\std{.00} & 0.91\std{.01} & 0.90\std{.00} & 0.92\std{.02} & 0.90\std{.02} & 0.90\std{.00} \\
        SNAPS & 0.92\std{.01} & 0.93\std{.01} & 0.92\std{.01} & 0.97\std{.01} & 0.91\std{.01} & 0.91\std{.01} & - & 0.90\std{.00} & 0.90\std{.00} & 0.90\std{.00} & 0.90\std{.01} & 0.90\std{.01} & 0.90\std{.00} \\
        RR-Mondrian & 0.92\std{.01} & 0.91\std{.01} & 0.90\std{.01} & 0.90\std{.01} & 0.90\std{.01} & 0.90\std{.01} & 0.91\std{.01} & 0.90\std{.00} & 0.91\std{.01} & \textcolor{red}{0.89\std{.00}} & 0.92\std{.01} & 0.92\std{.02} & 0.90\std{.00} \\
        CalLCP & 0.94\std{.01} & 0.94\std{.01} & 0.94\std{.00} & 0.94\std{.01} & 0.94\std{.00} & 0.94\std{.00} & 0.94\std{.00} & OOT & 0.91\std{.00} & OOT & 0.95\std{.01} & 0.93\std{.01} & 0.92\std{.00} \\
        \midrule
        \ourmethod & 0.90\std{.01} & 0.92\std{.01} & 0.90\std{.00} & 0.90\std{.00} & 0.90\std{.00} & 0.90\std{.01} & 0.90\std{.01} & 0.90\std{.00} & 0.90\std{.00} & 0.90\std{.00} & 0.90\std{.01} & 0.90\std{.00} & 0.90\std{.00} \\
        \bottomrule
    \end{tabular}}
    \label{tab:marginal_coverage_all}
\end{table*}

\subsection{Ablation Studies}

Our method achieves robust conditional coverage by replacing
embedding-space proximity with a structure-aware PPR kernel.
To isolate the effect of structural localization from
embedding-space reconditioning during densification, we replace
both PPR-based anchor sampling and calibration weighting with
Gaussian localization (GSS). We sample a continuous anchor from
a Gaussian centered at the test-node representation and weight
calibration and test nodes by their similarities to this anchor.
We preserve the same densification and preprocessing settings,
including PCA and eigenvalue-based bandwidths when enabled.
Table~\ref{tab:ppr_vs_gss} shows that PPR consistently improves
WSC at the cost of larger prediction sets, highlighting the
benefit of multi-hop structural localization over direct
embedding-space similarities.

\begin{table*}[t]
\scriptsize
\centering
\caption{Isolated kernel type effect. A PPR kernel provides better coverage but a larger prediction interval/set than a Gaussian (GSS) kernel.}
\vspace{-1mm}
\resizebox{\textwidth}{!}{
    \begin{tabular}{L{0.85in}L{0.6in}C{0.3in}C{0.3in}C{0.3in}C{0.3in}C{0.3in}C{0.3in}C{0.3in}C{0.3in}C{0.3in}C{0.3in}}
        \toprule
        \textbf{Measure} & \textbf{Variant} & \textbf{ANH} & \textbf{CHI} & \textbf{EDU} & \textbf{ELC} & \textbf{INC} & \textbf{UNM} & \textbf{TWT} & \textbf{CRA} & \textbf{CBAS} & \textbf{WKB} \\
        \midrule
        \multirow{2}{*}{Coverage} & GSS & 0.63\std{.04} & 0.62\std{.02} & 0.70\std{.01} & 0.59\std{.02} & 0.69\std{.02} & 0.72\std{.02} & 0.71\std{.02} & 0.69\std{.02} & 0.52\std{.06} & 0.66\std{.02} \\
         & PPR & 0.68\std{.03} & 0.71\std{.02} & 0.75\std{.01} & 0.68\std{.01} & 0.75\std{.01} & 0.77\std{.01} & 0.73\std{.03} & 0.72\std{.01} & 0.63\std{.02} & 0.70\std{.01} \\
        \midrule
        \multirow{2}{*}{Prediction Length} & GSS & 1.60\std{.09} & 2.00\std{.09} & 2.21\std{.15} & 1.06\std{.04} & 2.14\std{.11} & 1.79\std{.06} & 2.02\std{.11} & 0.90\std{.05} & 2.04\std{.10} & 3.60\std{.14} \\
         & PPR & 1.66\std{.12} & 2.17\std{.14} & 2.44\std{.24} & 1.11\std{.06} & 2.52\std{.21} & 1.85\std{.05} & 2.10\std{.17} & 2.00\std{.04} & 2.75\std{.09} & 4.21\std{.06} \\
        \bottomrule
    \end{tabular}}
    \label{tab:ppr_vs_gss}
\end{table*}

\begin{table*}[h]
\scriptsize
\centering
\caption{Ablation study on the main components across regression and classification with $\alpha=0.1$. Results show that the most significant component is densification particularly for regression. }
\vspace{-1mm}
\resizebox{\textwidth}{!}{
    \begin{tabular}{L{0.7in}C{0.25in}C{0.25in}C{0.25in}C{0.25in}C{0.25in}C{0.25in}C{0.25in}C{0.25in}C{0.25in}C{0.25in}C{0.25in}C{0.25in}C{0.25in}C{0.25in}C{0.25in}C{0.25in}C{0.25in}C{0.25in}}
        \toprule
        \textbf{Variant} & \multicolumn{2}{c}{\textbf{ANH}} & \multicolumn{2}{c}{\textbf{CHI}} & \multicolumn{2}{c}{\textbf{TWT}} & \multicolumn{2}{c}{\textbf{EDU}} & \multicolumn{2}{c}{\textbf{ELC}} & \multicolumn{2}{c}{\textbf{INC}} & \multicolumn{2}{c}{\textbf{UNM}} & \multicolumn{2}{c}{\textbf{CBAS}} & \multicolumn{2}{c}{\textbf{WKB}} \\
        \cmidrule(lr){2-3} \cmidrule(lr){4-5} \cmidrule(lr){6-7} \cmidrule(lr){8-9} \cmidrule(lr){10-11} \cmidrule(lr){12-13} \cmidrule(lr){14-15} \cmidrule(lr){16-17} \cmidrule(lr){18-19}
         & \textbf{Cov} & \textbf{Size} & \textbf{Cov} & \textbf{Size} & \textbf{Cov} & \textbf{Size} & \textbf{Cov} & \textbf{Size} & \textbf{Cov} & \textbf{Size} & \textbf{Cov} & \textbf{Size} & \textbf{Cov} & \textbf{Size} & \textbf{Cov} & \textbf{Size} & \textbf{Cov} & \textbf{Size} \\
        \midrule
        \ourmethod & 0.68\std{.03} & 1.66\std{.12} & 0.71\std{.02} & 2.17\std{.14} & 0.73\std{.03} & 2.10\std{.17} & 0.75\std{.01} & 2.44\std{.24} & 0.68\std{.01} & 1.11\std{.06} & 0.75\std{.01} & 2.52\std{.21} & 0.77\std{.01} & 1.85\std{.05} & 0.63\std{.02} & 2.75\std{.09} & 0.70\std{.01} & 4.21\std{.06} \\
        \midrule
        w/o Densification & 0.46\std{.01} & 1.14\std{.09} & 0.52\std{.02} & 1.80\std{.08} & 0.63\std{.03} & 1.96\std{.11} & 0.72\std{.01} & 2.61\std{.38} & 0.70\std{.02} & 1.32\std{.17} & 0.70\std{.01} & 2.67\std{.54} & 0.74\std{.01} & 2.00\std{.20} & 0.62\std{.02} & 2.88\std{.07} & 0.69\std{.02} & 4.14\std{.04} \\
        w/o PCA & 0.67\std{.02} & 1.64\std{.13} & 0.62\std{.02} & 1.94\std{.04} & 0.73\std{.02} & 2.19\std{.13} & 0.74\std{.01} & 2.44\std{.19} & 0.65\std{.01} & 1.05\std{.04} & 0.72\std{.02} & 2.35\std{.14} & 0.76\std{.01} & 1.80\std{.05} & 0.63\std{.02} & 2.75\std{.09} & 0.67\std{.01} & 4.11\std{.05} \\
        \bottomrule
    \end{tabular}}
    \label{tab:ablationwsc}
\end{table*}

%\subsubsection{Add-on Effects}

\textbf{Add-on Effects.} While \ourmethod relies heavily on the structure due to the PPR-based kernel, the densification step, which incorporates useful information from the embedding space, plays an important role. Table \ref{tab:ablationwsc} shows that densification often has the largest effect, particularly on regression datasets (ANH, CHI, and TWT), although the gains are not uniform. PCA offers a more modest benefit by reducing redundant embedding dimensions. In practice, we apply PCA when the low-dimensional space is more useful for densification (see Appendix \ref{app:experiments}).
\section{Conclusion}
\label{sec:conclusion}
\vspace{-2mm}
We have introduced \ourmethod, a structure-aware localized conformal prediction framework for \gnns that explicitly incorporates graph topology into conformal localization and weighting. By replacing embedding-only proximity with a PPR-based kernel, our method achieves reliable marginal coverage while improving worst-slab coverage. Experiments on regression and classification datasets demonstrate the benefits of structure-aware localization, with a prediction-length trade-off. We believe this work highlights the importance of structure-aware localization for UQ on graphs. An important future direction is to extend localized conformal prediction for dynamic graphs.\\
\noindent
\textbf{Limitations.}
PPR-based localization can be less effective on sparse graphs.
While densification mitigates this issue, poor embeddings may
introduce noisy edges or miss informative connections depending
on the similarity threshold. Longer PPR walks may weaken
localization, while small calibration sets can lead to unstable
weights. Moreover, improved conditional coverage may require
larger prediction sets, and densification and PPR computations
introduce additional runtime costs, particularly on large graphs.

% \textbf{Limitations. }The effectiveness of the PPR kernel depends on the sparsity of the graph. The introduced densification step could remedy the effect of strong localization. Extremely poor embeddings, however, can lead to the addition of noisy edges at low thresholds or to the creation of rare, useless connections at high thresholds. Besides, the PPR with long random walks allows more nodes to contribute and reduces the sampling remaining probability, while potentially increasing computational costs on densely connected large graphs. Finally, we forcefully sample anchor nodes from the calibration nodes, which can increase both bias and variance errors and lower generalization on very small calibration sets.
%\newpage
\medskip
{
\small
\bibliographystyle{plain}
\bibliography{graphlcp}

@article{shafer2008tutorial,
  title={A tutorial on conformal prediction.},
  author={Shafer, Glenn and Vovk, Vladimir},
  journal={Journal of Machine Learning Research},
  volume={9},
  number={3},
  year={2008}
}

@article{angelopoulos2021gentle,
  title={A gentle introduction to conformal prediction and distribution-free uncertainty quantification},
  author={Angelopoulos, Anastasios N and Bates, Stephen},
  journal={arXiv preprint arXiv:2107.07511},
  year={2021}
}

@inproceedings{mpnn2017,
author = {Gilmer, Justin and Schoenholz, Samuel S. and Riley, Patrick F. and Vinyals, Oriol and Dahl, George E.},
title = {Neural message passing for Quantum chemistry},
year = {2017},
publisher = {JMLR.org},
booktitle = {Proceedings of the 34th International Conference on Machine Learning - Volume 70},
pages = {1263–1272},
numpages = {10},
location = {Sydney, NSW, Australia},
series = {ICML'17}
}

@book{maincp2005,
title = "Algorithmic Learning in a Random World",
author = "Vladimir Vovk and Alex Gammerman and Glenn Shafer",
note = "Springer, New York",
year = "2005",
language = "English",
publisher = "Springer",
}

@article{infeasiblecondcp2020,
    author = {Foygel Barber, Rina and Candès, Emmanuel J and Ramdas, Aaditya and Tibshirani, Ryan J},
    title = {The limits of distribution-free conditional predictive inference},
    journal = {Information and Inference: A Journal of the IMA},
    volume = {10},
    number = {2},
    pages = {455-482},
    year = {2020},
    month = {08},
    issn = {2049-8772},
    doi = {10.1093/imaiai/iaaa017},
    url = {https://doi.org/10.1093/imaiai/iaaa017},
    eprint = {https://academic.oup.com/imaiai/article-pdf/10/2/455/38549621/iaaa017.pdf},
}

@article{rlcp2024,
    author = {Hore, Rohan and Barber, Rina Foygel},
    title = {Conformal prediction with local weights: randomization enables robust guarantees},
    journal = {Journal of the Royal Statistical Society Series B: Statistical Methodology},
    volume = {87},
    number = {2},
    pages = {549-578},
    year = {2024},
    month = {11},
    issn = {1369-7412},
    doi = {10.1093/jrsssb/qkae103},
    url = {https://doi.org/10.1093/jrsssb/qkae103},
    eprint = {https://academic.oup.com/jrsssb/article-pdf/87/2/549/60583919/qkae103.pdf},
}

@inproceedings{wcp2019,
 author = {Tibshirani, Ryan J and Foygel Barber, Rina and Candes, Emmanuel and Ramdas, Aaditya},
 booktitle = {Advances in Neural Information Processing Systems},
 editor = {H. Wallach and H. Larochelle and A. Beygelzimer and F. d\textquotesingle Alch\'{e}-Buc and E. Fox and R. Garnett},
 pages = {},
 publisher = {Curran Associates, Inc.},
 title = {Conformal Prediction Under Covariate Shift},
 url = {https://proceedings.neurips.cc/paper_files/paper/2019/file/8fb21ee7a2207526da55a679f0332de2-Paper.pdf},
 volume = {32},
 year = {2019}
}

@article{cfgnn2023,
  title={Uncertainty quantification over graph with conformalized graph neural networks},
  author={Huang, Kexin and Jin, Ying and Candes, Emmanuel and Leskovec, Jure},
  journal={NeurIPS},
  year={2023}
}

@inproceedings{cpforgnn2023,
  title={Conformal prediction sets for graph neural networks},
  author={Zargarbashi, Soroush H and Antonelli, Simone and Bojchevski, Aleksandar},
  booktitle={International Conference on Machine Learning},
  pages={12292--12318},
  year={2023},
  organization={PMLR}
}

@inproceedings{
cagcn2021,
title={Be Confident! Towards Trustworthy Graph Neural Networks via Confidence Calibration},
author={Xiao Wang and Hongrui Liu and Chuan Shi and Cheng Yang},
booktitle={Advances in Neural Information Processing Systems},
editor={A. Beygelzimer and Y. Dauphin and P. Liang and J. Wortman Vaughan},
year={2021},
url={https://openreview.net/forum?id=9c-IsSptbmA}
}

@inproceedings{gats2022,
 author = {Hsu, Hans Hao-Hsun and Shen, Yuesong and Tomani, Christian and Cremers, Daniel},
 booktitle = {Advances in Neural Information Processing Systems},
 editor = {S. Koyejo and S. Mohamed and A. Agarwal and D. Belgrave and K. Cho and A. Oh},
 pages = {13775--13786},
 publisher = {Curran Associates, Inc.},
 title = {What Makes Graph Neural Networks Miscalibrated?},
 volume = {35},
 year = {2022}
}

@inproceedings{naps2023,
author = {Clarkson, Jase},
title = {Distribution free prediction sets for node classification},
year = {2023},
publisher = {JMLR.org},
booktitle = {Proceedings of the 40th International Conference on Machine Learning},
articleno = {249},
numpages = {11},
location = {Honolulu, Hawaii, USA},
series = {ICML'23}
}

@misc{networkcp2023,
      title={On the Validity of Conformal Prediction for Network Data Under Non-Uniform Sampling}, 
      author={Robert Lunde},
      year={2023},
      eprint={2306.07252},
      archivePrefix={arXiv},
      primaryClass={math.ST},
      url={https://arxiv.org/abs/2306.07252}, 
}

@misc{ranklist2021,
      title={Exchangeability, Conformal Prediction, and Rank Tests}, 
      author={Arun Kumar Kuchibhotla},
      year={2021},
      eprint={2005.06095},
      archivePrefix={arXiv},
      primaryClass={stat.ME},
      url={https://arxiv.org/abs/2005.06095}, 
}

@article{networkassistedcp2025,
author = {Robert Lunde and Elizaveta Levina and Ji Zhu},
title = {Conformal Prediction for Network-Assisted Regression},
journal = {Journal of the American Statistical Association},
volume = {120},
number = {551},
pages = {1633--1644},
year = {2025},
publisher = {Taylor \& Francis},
doi = {10.1080/01621459.2025.2506198},
URL = {https://doi.org/10.1080/01621459.2025.2506198},
eprint = {https://doi.org/10.1080/01621459.2025.2506198}
}

@article{linkfdr2023,
  title={Conformal link prediction for false discovery rate control},
  author={Ariane Marandon},
  journal={TEST},
  year={2023},
  volume={33},
  pages={1062 - 1083},
  url={https://api.semanticscholar.org/CorpusID:259252560}
}

@ARTICLE{skipconnection2023,
  author={Oyedotun, Oyebade K. and Ismaeil, Kassem Al and Aouada, Djamila},
  journal={IEEE Transactions on Neural Networks and Learning Systems}, 
  title={Why Is Everyone Training Very Deep Neural Network With Skip Connections?}, 
  year={2023},
  volume={34},
  number={9},
  pages={5961-5975},
  doi={10.1109/TNNLS.2021.3131813}}

@article{trustyourneighbor2021,
  title={Trust your neighbors: A comprehensive survey of neighborhood-based methods for recommender systems},
  author={Athanasios N. Nikolakopoulos and Xia Ning and Christian Desrosiers and George Karypis},
  journal={ArXiv},
  year={2021},
  volume={abs/2109.04584},
  url={https://api.semanticscholar.org/CorpusID:237485190}
}

@inproceedings{ppr2003,
author = {Jeh, Glen and Widom, Jennifer},
title = {Scaling personalized web search},
year = {2003},
isbn = {1581136803},
publisher = {Association for Computing Machinery},
address = {New York, NY, USA},
url = {https://doi.org/10.1145/775152.775191},
doi = {10.1145/775152.775191},
booktitle = {Proceedings of the 12th International Conference on World Wide Web},
pages = {271–279},
numpages = {9},
location = {Budapest, Hungary},
series = {WWW '03}
}

@inproceedings{
appnp2019,
title={Combining Neural Networks with Personalized PageRank for Classification on Graphs},
author={Johannes Gasteiger and Aleksandar Bojchevski and Stephan Günnemann},
booktitle={International Conference on Learning Representations},
year={2019},
url={https://openreview.net/forum?id=H1gL-2A9Ym},
}

@article{
oversquashingeffect2024,
title={How does over-squashing affect the power of {GNN}s?},
author={Francesco Di Giovanni and T. Konstantin Rusch and Michael Bronstein and Andreea Deac and Marc Lackenby and Siddhartha Mishra and Petar Veli{\v{c}}kovi{\'c}},
journal={Transactions on Machine Learning Research},
issn={2835-8856},
year={2024},
url={https://openreview.net/forum?id=KJRoQvRWNs},
note={}
}

@inproceedings{oversmoothingoversquashingtradeoff2023,
author = {Giraldo, Jhony H. and Skianis, Konstantinos and Bouwmans, Thierry and Malliaros, Fragkiskos D.},
title = {On the Trade-off between Over-smoothing and Over-squashing in Deep Graph Neural Networks},
year = {2023},
isbn = {9798400701245},
publisher = {Association for Computing Machinery},
address = {New York, NY, USA},
url = {https://doi.org/10.1145/3583780.3614997},
doi = {10.1145/3583780.3614997},
booktitle = {Proceedings of the 32nd ACM International Conference on Information and Knowledge Management},
pages = {566–576},
numpages = {11},
location = {Birmingham, United Kingdom},
series = {CIKM '23}
}

@inproceedings{
resisitoversmoothing2024,
title={Resisting Over-Smoothing in Graph Neural Networks via Dual-Dimensional Decoupling},
author={Wei Shen and Mang Ye and Wenke Huang},
booktitle={ACM Multimedia 2024},
year={2024},
url={https://openreview.net/forum?id=QqG3wI3c8L}
}

@misc{oversmoothingsurvey2023,
      title={A Survey on Oversmoothing in Graph Neural Networks}, 
      author={T. Konstantin Rusch and Michael M. Bronstein and Siddhartha Mishra},
      year={2023},
      eprint={2303.10993},
      archivePrefix={arXiv},
      primaryClass={cs.LG},
      url={https://arxiv.org/abs/2303.10993}, 
}

@inproceedings{rrgnn2025,
author = {Zhang, Zheng and Bao, Jie and Zhou, Zhixin and Colombo, Nicolo and Cheng, Lixin and Luo, Rui},
title = {Residual reweighted conformal prediction for graph neural networks},
year = {2025},
publisher = {JMLR.org},
booktitle = {Proceedings of the Forty-First Conference on Uncertainty in Artificial Intelligence},
articleno = {223},
numpages = {18},
location = {Rio de Janeiro, Brazil},
series = {UAI '25}
}

@inproceedings{stronglocalizationppr2015,
author = {Nassar, Huda and Kloster, Kyle and Gleich, David F.},
title = {Strong Localization in Personalized PageRank Vectors},
year = {2015},
isbn = {9783319267838},
publisher = {Springer-Verlag},
address = {Berlin, Heidelberg},
url = {https://doi.org/10.1007/978-3-319-26784-5_15},
doi = {10.1007/978-3-319-26784-5_15},
booktitle = {Proceedings of the 12th International Workshop on Algorithms and Models for the Web Graph - Volume 9479},
pages = {190–202},
numpages = {13},
location = {Eindhoven, The Netherlands},
series = {WAW 2015}
}

@misc{pca2014,
      title={A Tutorial on Principal Component Analysis}, 
      author={Jonathon Shlens},
      year={2014},
      eprint={1404.1100},
      archivePrefix={arXiv},
      primaryClass={cs.LG},
      url={https://arxiv.org/abs/1404.1100}, 
}

@inproceedings{residualgnn2020,
author = {Jia, Junteng and Benson, Austion R.},
title = {Residual Correlation in Graph Neural Network Regression},
year = {2020},
isbn = {9781450379984},
publisher = {Association for Computing Machinery},
address = {New York, NY, USA},
url = {https://doi.org/10.1145/3394486.3403101},
doi = {10.1145/3394486.3403101},
booktitle = {Proceedings of the 26th ACM SIGKDD International Conference on Knowledge Discovery \& Data Mining},
pages = {588–598},
numpages = {11},
location = {Virtual Event, CA, USA},
series = {KDD '20}
}

@misc{conditionalcp2025,
      title={Conditional Coverage Diagnostics for Conformal Prediction}, 
      author={Sacha Braun and David Holzmüller and Michael I. Jordan and Francis Bach},
      year={2025},
      eprint={2512.11779},
      archivePrefix={arXiv},
      primaryClass={stat.ML},
      url={https://arxiv.org/abs/2512.11779}, 
}

@article{callcp2022,
    author = {Guan, Leying},
    title = {Localized conformal prediction: a generalized inference framework for conformal prediction},
    journal = {Biometrika},
    volume = {110},
    number = {1},
    pages = {33-50},
    year = {2022},
    month = {07},
    issn = {1464-3510},
    doi = {10.1093/biomet/asac040},
    url = {https://doi.org/10.1093/biomet/asac040},
    eprint = {https://academic.oup.com/biomet/article-pdf/110/1/33/49160126/asac040.pdf},
}

@article{ma2023histgnn,
  title={HiSTGNN: Hierarchical spatio-temporal graph neural network for weather forecasting},
  author={Ma, Minbo and Xie, Peng and Teng, Fei and Wang, Bin and Ji, Shenggong and Zhang, Junbo and Li, Tianrui},
  journal={Information Sciences},
  volume={648},
  pages={119580},
  year={2023},
  publisher={Elsevier}
}

@article{cheng2025graph,
  title={Graph neural networks for financial fraud detection: a review},
  author={Cheng, Dawei and Zou, Yao and Xiang, Sheng and Jiang, Changjun},
  journal={Frontiers of Computer Science},
  volume={19},
  number={9},
  pages={199609},
  year={2025},
  publisher={Springer}
}

@article{feinberg2018potential,
  title={PotentialNet for molecular property prediction},
  author={Feinberg, Evan N and Sur, Debnil and Wu, Zhenqin and Husic, Brooke E and Mai, Huanghao and Li, Yang and Sun, Saisai and Yang, Jianyi and Ramsundar, Bharath and Pande, Vijay S},
  journal={ACS central science},
  volume={4},
  number={11},
  pages={1520},
  year={2018}
}

@article{beyondexchangeability2022,
  title={Conformal prediction beyond exchangeability},
  author={Rina Foygel Barber and Emmanuel J. Cand{\`e}s and Aaditya Ramdas and Ryan J. Tibshirani},
  journal={The Annals of Statistics},
  year={2022},
  url={https://api.semanticscholar.org/CorpusID:247158820}
}

@inproceedings{
highdimensionalcovarite2025,
title={Conformal Inference under High-Dimensional Covariate Shifts via Likelihood-Ratio Regularization},
author={Sunay Joshi and Shayan Kiyani and George J. Pappas and Edgar Dobriban and Hamed Hassani},
booktitle={The Thirty-ninth Annual Conference on Neural Information Processing Systems},
year={2025},
url={https://openreview.net/forum?id=2bgwni6Ber}
}

@article{validconfidenseset2021,
author = {Cauchois, Maxime and Gupta, Suyash and Duchi, John C.},
title = {Knowing what you know: valid and validated confidence sets in multiclass and multilabel prediction},
year = {2021},
issue_date = {January 2021},
publisher = {JMLR.org},
volume = {22},
number = {1},
issn = {1532-4435},
journal = {J. Mach. Learn. Res.},
month = jan,
articleno = {81},
numpages = {42}
}

@inproceedings{validadaptiveacoverage2020,
author = {Romano, Yaniv and Sesia, Matteo and Cand\`{e}s, Emmanuel J.},
title = {Classification with valid and adaptive coverage},
year = {2020},
isbn = {9781713829546},
publisher = {Curran Associates Inc.},
address = {Red Hook, NY, USA},
booktitle = {Proceedings of the 34th International Conference on Neural Information Processing Systems},
articleno = {302},
numpages = {11},
location = {Vancouver, BC, Canada},
series = {NIPS '20}
}

@inproceedings{codrug2023,
author = {Laghuvarapu, Siddhartha and Lin, Zhen and Sun, Jimeng},
title = {CoDrug: conformai drug property prediction with density estimation under covariate shift},
year = {2023},
publisher = {Curran Associates Inc.},
address = {Red Hook, NY, USA},
booktitle = {Proceedings of the 37th International Conference on Neural Information Processing Systems},
articleno = {1641},
numpages = {20},
location = {New Orleans, LA, USA},
series = {NIPS '23}
}

@inproceedings{cplinprediction2024,
author = {Zhao, Tianyi and Kang, Jian and Cheng, Lu},
title = {Conformalized Link Prediction on Graph Neural Networks},
year = {2024},
isbn = {9798400704901},
publisher = {Association for Computing Machinery},
address = {New York, NY, USA},
url = {https://doi.org/10.1145/3637528.3672061},
doi = {10.1145/3637528.3672061},
booktitle = {Proceedings of the 30th ACM SIGKDD Conference on Knowledge Discovery and Data Mining},
pages = {4490–4499},
numpages = {10},
location = {Barcelona, Spain},
series = {KDD '24}
}

@inproceedings{
inductivecpgnn2024,
title={Conformal Inductive Graph Neural Networks},
author={Soroush H. Zargarbashi and Aleksandar Bojchevski},
booktitle={The Twelfth International Conference on Learning Representations},
year={2024},
url={https://openreview.net/forum?id=homn1jOKI5}
}

@inproceedings{oodcpgnn2025,
author = {Lin, Xixun and Cao, Yanan and Sun, Nan and Zou, Lixin and Zhou, Chuan and Zhang, Peng and Zhang, Shuai and Zhang, Ge and Wu, Jia},
title = {Conformal Graph-level Out-of-distribution Detection with Adaptive Data Augmentation},
year = {2025},
isbn = {9798400712746},
publisher = {Association for Computing Machinery},
address = {New York, NY, USA},
url = {https://doi.org/10.1145/3696410.3714879},
doi = {10.1145/3696410.3714879},
booktitle = {Proceedings of the ACM on Web Conference 2025},
pages = {4755–4765},
numpages = {11},
location = {Sydney NSW, Australia},
series = {WWW '25}
}

@inproceedings{
citationful2018,
title={Deep Gaussian Embedding of Graphs:  Unsupervised Inductive Learning via Ranking},
author={Aleksandar Bojchevski and Stephan Günnemann},
booktitle={International Conference on Learning Representations},
year={2018},
url={https://openreview.net/forum?id=r1ZdKJ-0W},
}

@misc{coauthor2019,
      title={Pitfalls of Graph Neural Network Evaluation}, 
      author={Oleksandr Shchur and Maximilian Mumme and Aleksandar Bojchevski and Stephan Günnemann},
      year={2019},
      eprint={1811.05868},
      archivePrefix={arXiv},
      primaryClass={cs.LG},
      url={https://arxiv.org/abs/1811.05868}, 
}

@inproceedings{
good2022,
title={{GOOD}: A Graph Out-of-Distribution Benchmark},
author={Shurui Gui and Xiner Li and Limei Wang and Shuiwang Ji},
booktitle={Thirty-sixth Conference on Neural Information Processing Systems Datasets and Benchmarks Track},
year={2022},
url={https://openreview.net/forum?id=8hHg-zs_p-h}
}

@inproceedings{
gcn2017,
title={Semi-Supervised Classification with Graph Convolutional Networks},
author={Thomas N. Kipf and Max Welling},
booktitle={International Conference on Learning Representations},
year={2017},
url={https://openreview.net/forum?id=SJU4ayYgl}
}

@inproceedings{
snaps2024,
title={Similarity-Navigated Conformal Prediction for Graph Neural Networks},
author={Jianqing Song and Jianguo Huang and Wenyu Jiang and Baoming Zhang and Shuangjie Li and Chongjun Wang},
booktitle={The Thirty-eighth Annual Conference on Neural Information Processing Systems},
year={2024},
url={https://openreview.net/forum?id=iBZSOh027z}
}

@article{discreteisoperimetric1996,
  title={Discrete isoperimetric inequalities},
  author={Chung Fan},
  journal={Surveys in differential geometry},
  year={1996},
  volume={9},
  pages={53-82},
  url={https://api.semanticscholar.org/CorpusID:123668239}
}

@article{onvanishinggradients2026,
  title={On vanishing gradients, over-smoothing, and over-squashing in gnns: Bridging recurrent and graph learning},
  author={Arroyo, {\'A}lvaro and Gravina, Alessio and Gutteridge, Benjamin and Barbero, Federico and Gallicchio, Claudio and Dong, Xiaowen and Bronstein, Michael and Vandergheynst, Pierre},
  journal={Advances in Neural Information Processing Systems},
  volume={38},
  pages={74356--74393},
  year={2026}
}

@inproceedings{cqr2019,
 author = {Romano, Yaniv and Patterson, Evan and Candes, Emmanuel},
 booktitle = {Advances in Neural Information Processing Systems},
 editor = {H. Wallach and H. Larochelle and A. Beygelzimer and F. d\textquotesingle Alch\'{e}-Buc and E. Fox and R. Garnett},
 pages = {},
 publisher = {Curran Associates, Inc.},
 title = {Conformalized Quantile Regression},
 url = {https://proceedings.neurips.cc/paper_files/paper/2019/file/5103c3584b063c431bd1268e9b5e76fb-Paper.pdf},
 volume = {32},
 year = {2019}
}
%%%%%%%%%%%%%%%%%%%%%%%%%%%%%%%%%%%%%%%%%%%%%%%%%%%%%%%%%%%%
}
\newpage
\appendix
\section{Appendix}

\subsection{Proofs and further boundary results}
\label{app:boundary_proofs}
The graph throughout these proofs is finite, connected, simple,
undirected, and unweighted, with $\varnothing\neq B\subsetneq V$.
All distances in a disconnected induced subgraph are taken to be $+\infty$.

\subsubsection{Embedding-space boundary result}
\label{app:embedding_boundary}

\begin{lem}[Boundary reduction to cross-set distances]
\label{lem:embedding_cross_set}
Let $U$ and $W$ be independent and uniformly distributed on
$B$ and $B^c$, respectively, and define
$$
Q_B(Z)
:=
\mathbb E\!\left[\|z_U-z_W\|_2^2\right]
=
\frac{1}{|B||B^c|}
\sum_{u\in B}\sum_{v\in B^c}\|z_u-z_v\|_2^2.
$$
For every $\delta>0$,
\begin{equation}
\frac{|\operatorname{bd}_{Z}(B;2\delta)|}{|B|}
\geq
\left[1-\frac{Q_B(Z)}{4\delta^2}\right]_+.
\label{eq:embedding_cross_set_bound}
\end{equation}
\end{lem}

\begin{proof}

Fix $u\in B$. By definition,
$u\in\operatorname{bd}_{Z}(B;2\delta)$ if and only if there is
at least one $v\in B^c$ such that $\|z_u-z_v\|_2\leq2\delta$.
Hence,
$$
\mathbf 1\{u\in\operatorname{bd}_{Z}(B;2\delta)\}
\geq
\frac{1}{|B^c|}
\sum_{v\in B^c}
\mathbf 1\{\|z_u-z_v\|_2\leq2\delta\}.
$$
Indeed, if no such $v$ exists, both sides are zero. Otherwise,
the left-hand side is one and the right-hand side is a fraction
between zero and one. Therefore we have:
$$
\frac{1}{|B|}
\sum_{u\in B}
\mathbf 1\{u\in\operatorname{bd}_{Z}(B;2\delta)\}
\geq
\frac{1}{|B||B^c|}
\sum_{u\in B}\sum_{v\in B^c}
\mathbf 1\{\|z_u-z_v\|_2\leq2\delta\}.
$$
The left-hand side counts the vertices in the boundary and equals
$|\operatorname{bd}_{Z}(B;2\delta)|/|B|$.
Because $U$ and $W$ are independent and uniform on $B$ and $B^c$,
respectively, each pair $(u,v)\in B\times B^c$ has probability
$1/(|B||B^c|)$. Thus the right-hand side is
$\mathbb P(\|z_U-z_W\|_2\leq2\delta)$, and
$$
\frac{|\operatorname{bd}_{Z}(B;2\delta)|}{|B|}
\geq
\mathbb P(\|z_U-z_W\|_2\leq 2\delta)
=
1-\mathbb P(\|z_U-z_W\|_2>2\delta).
$$

By Markov's inequality,
$$
\mathbb P(\|z_U-z_W\|_2>2\delta)
\leq
\frac{\mathbb E[\|z_U-z_W\|_2^2]}{4\delta^2}
=
\frac{Q_B(Z)}{4\delta^2}.
$$
The boundary fraction is also nonnegative. Combining these facts
proves \eqref{eq:embedding_cross_set_bound}.

\end{proof}

\begin{lem}[Vector-valued Poincaré inequality]
\label{lem:embedding_vector_poincare}
Let $Z=(z_u)_{u\in V}\in\mathbb R^{N\times p}$ and
$\bar z=N^{-1}\sum_{u\in V}z_u$. Then
\begin{equation}
\sigma_1\sum_{u\in V}\|z_u-\bar z\|_2^2
\leq
\mathcal E_G(Z).
\label{eq:embedding_vector_poincare}
\end{equation}
\end{lem}

\begin{proof}
The variational characterization of the smallest nonzero
eigenvalue of the combinatorial Laplacian $L_G=D-A$ gives
\begin{equation}
\sigma_1
=
\min_{\substack{f:V\to\mathbb R,\ f\ne0\\
                \sum_{u\in V}f(u)=0}}
\frac{\sum_{\{u,v\}\in E}(f(u)-f(v))^2}
     {\sum_{u\in V}f(u)^2}
\label{eq:chung_rayleigh}
\end{equation}
This is the Rayleigh--Ritz formula for $L_G$ given by
Chung~\cite{discreteisoperimetric1996}. For each coordinate
$j$, set $f_j(u)=z_{u,j}-\bar z_j$.
Then $\sum_{u\in V}f_j(u)=0$, so
\eqref{eq:chung_rayleigh} implies
$$
\sigma_1\sum_{u\in V}(z_{u,j}-\bar z_j)^2
\leq
\sum_{\{u,v\}\in E}(z_{u,j}-z_{v,j})^2.
$$
The inequality also holds if $f_j$ is identically zero.
Summing over $j=1,\ldots,p$ proves the claim.
\end{proof}

\begin{lem}[Spectral control of cross-set dispersion]
\label{lem:embedding_poincare}
For a connected graph $G$ and $\rho_B=|B|/N$,
\begin{equation}
Q_B(Z)
\leq
\frac{N}{|B||B^c|\sigma_1}\mathcal E_G(Z)
=
\frac{\mathcal E_G(Z)}
{\rho_B(1-\rho_B)N\sigma_1}.
\label{eq:embedding_cross_set_poincare}
\end{equation}
\end{lem}

\begin{proof}

Let $\bar z=N^{-1}\sum_{u\in V}z_u$. Expanding squared distances gives
$$
\begin{aligned}
\sum_{\{u,v\}\subseteq V}\|z_u-z_v\|_2^2
&=
\frac12\sum_{u,v\in V}
 \bigl(\|z_u\|_2^2+\|z_v\|_2^2-2z_u^\top z_v\bigr)\\
&=
N\sum_{u\in V}\|z_u\|_2^2
-\left\|\sum_{u\in V}z_u\right\|_2^2\\
&=
N\sum_{u\in V}\|z_u-\bar z\|_2^2\\
&\leq
\frac{N}{\sigma_1}\mathcal E_G(Z),
\end{aligned}
$$
where the last step uses Lemma~\ref{lem:embedding_vector_poincare}.
The pairs with $u\in B$ and $v\in B^c$ are a subset of all
unordered vertex pairs. Therefore,
$$
\sum_{u\in B}\sum_{v\in B^c}\|z_u-z_v\|_2^2
\leq
\frac{N}{\sigma_1}\mathcal E_G(Z).
$$
Dividing both sides by $|B||B^c|$ proves
\eqref{eq:embedding_cross_set_poincare}.
\end{proof}

\begin{proof}[Proof of Theorem~\ref{thm:embedding_boundary}]
We obtain \eqref{eq:embedding_boundary_bound} by substituting \eqref{eq:embedding_cross_set_poincare} into
\eqref{eq:embedding_cross_set_bound}.

For the final statement, the proof of
Lemma~\ref{lem:embedding_poincare} gives
$$
\sum_{\{u,v\}\subseteq V}\|z_u^{(K)}-z_v^{(K)}\|_2^2
\leq
\frac{N}{\sigma_1}\mathcal E_G(Z_K)
\longrightarrow 0.
$$
The sum contains finitely many nonnegative terms. Hence, for
all sufficiently large $K$, every pair of vertices satisfies
$\|z_u^{(K)}-z_v^{(K)}\|_2<2\delta$. Since $B^c\neq\varnothing$,
every $u\in B$ then belongs to
$\operatorname{bd}_{Z_K}(B;2\delta)$.
\end{proof}

The contractivity analysis of \cite{onvanishinggradients2026}
shows that contractive GNN layers lead to vanishing
Dirichlet energy. Combining this result with
Theorem~\ref{thm:embedding_boundary}, we obtain
the following finite-depth bound on the embedding-space
boundary.

\begin{cor}[Finite-depth bound for contractive layers]
\label{cor:embedding_contractivity}
Let $Z_0=\mathbf X\in\mathbb R^{N\times p_0}$ be the input
node-feature matrix, and let
$Z_k=F_k(Z_{k-1})\in\mathbb R^{N\times p_k}$ for
$k=1,\ldots,K$, where $F_k$ is the $k$th GNN layer.
Suppose $F_k$ is Lipschitz with respect to the Frobenius norm with constant $\ell_k$ and $F_k(0)=0$.
Set $\Lambda_K=\prod_{k=1}^K\ell_k^2$ and
$\bar d=2|E|/N$. Then, for every nonempty
$B\subsetneq V$ and every $\delta>0$,
\begin{equation}
\frac{|\operatorname{bd}_{Z_K}(B;2\delta)|}{|B|}
\geq
\left[
1-
\frac{\bar d\,\Lambda_K\|Z_0\|_F^2}
{4\delta^2\rho_B(1-\rho_B)\sigma_1}
\right]_+.
\label{eq:embedding_contractive_bound}
\end{equation}
If $\ell_k\leq q$ for every $k$, where $0\leq q<1$,
then, on a fixed graph,
$\operatorname{bd}_{Z_K}(B;2\delta)=B$ for all
sufficiently large $K$.
\end{cor}

\begin{proof}
Since $F_k(0)=0$, the Lipschitz assumption gives
$$
\|Z_k\|_F
=
\|F_k(Z_{k-1})-F_k(0)\|_F
\leq
\ell_k\|Z_{k-1}\|_F.
$$
Applying this inequality successively yields
\begin{equation}
\|Z_K\|_F^2
\leq
\left(\prod_{k=1}^K\ell_k^2\right)\|Z_0\|_F^2
=
\Lambda_K\|Z_0\|_F^2.
\label{eq:embedding_contractive_norm}
\end{equation}

For any edge $\{u,v\}\in E$,
$$
\|z_u^{(K)}-z_v^{(K)}\|_2^2
\leq
2\bigl(\|z_u^{(K)}\|_2^2+
        \|z_v^{(K)}\|_2^2\bigr)
\leq
2\|Z_K\|_F^2.
$$
The last inequality holds because $u$ and $v$ are
distinct rows of $Z_K$. Summing over the $|E|$ edges
and using \eqref{eq:embedding_contractive_norm}, we obtain
$$
\begin{aligned}
\mathcal E_G(Z_K)
&=
\sum_{\{u,v\}\in E}
\|z_u^{(K)}-z_v^{(K)}\|_2^2\\
&\leq
2|E|\|Z_K\|_F^2\\
&\leq
2|E|\Lambda_K\|Z_0\|_F^2\\
&=
N\bar d\,\Lambda_K\|Z_0\|_F^2.
\end{aligned}
$$
Substituting this bound into
\eqref{eq:embedding_boundary_bound} proves
\eqref{eq:embedding_contractive_bound}.

Finally, if $\ell_k\leq q<1$ for all $k$, then
$\Lambda_K\leq q^{2K}\to0$. Thus the energy bound
above implies $\mathcal E_G(Z_K)\to0$, and the
eventual equality follows from
Theorem~\ref{thm:embedding_boundary}.
\end{proof}

\textbf{Remark. }For intermediate depths with $\mathcal E_G(Z_K)>0$,
Eq.~\ref{eq:embedding_boundary_bound} equivalently gives
\begin{equation}
\frac{
|B\setminus\operatorname{bd}_{Z_K}(B;2\delta)|
}{|B|}
\leq
\min\left\{
1,\,
\frac{\mathcal E_G(Z_K)}
{4\delta^2\rho_B(1-\rho_B)N\sigma_1}
\right\}.
\end{equation}
Thus, whenever the normalized energy term is below one,
the bound limits the fraction of nodes in $B$ that remain
farther than $2\delta$ from $B^c$.
Lower energy yields a tighter bound, although it does
not necessarily imply a monotonic increase in the
actual boundary fraction across GNN depths.

\subsubsection{Graph-space boundary result}
\label{app:graph_boundary_proof}

Define the one-hop inner boundary of $B$ by
$$
\partial_{\mathrm{in}}B
:=
\{u\in B: \exists v\in B^c
              \text{ s.t. }\{u,v\}\in E\}.
$$

\begin{lem}[First-exit identity]
\label{lem:graph_first_exit}
For every $u\in B$,
$$
d_G(u,B^c)
=
1+d_{G[B]}(u,\partial_{\mathrm{in}}B).
$$
Consequently, for every integer $r\geq1$,
\begin{equation}
\operatorname{bd}_G(B;r)
=
\{u\in B:
d_{G[B]}(u,\partial_{\mathrm{in}}B)\leq r-1\}.
\label{eq:graph_boundary_representation}
\end{equation}
\end{lem}

\begin{proof}
Let $z\in\partial_{\mathrm{in}}B$ be reachable from $u$
by a shortest path in $G[B]$. Since $z\in\partial_{\mathrm{in}}B$ has a neighbor in $B^c$, a path from $u$ to $z$ in $G[B]$ extends to $B^c$, therefore
$$
d_G(u,B^c)
\leq
1+d_{G[B]}(u,\partial_{\mathrm{in}}B).
$$

Conversely, consider a shortest path from $u$ to $B^c$,
of length $k=d_G(u,B^c)$. All vertices before its final
vertex lie in $B$; otherwise the path would have reached
$B^c$ sooner. Its second-to-last vertex on the path belongs
to $\partial_{\mathrm{in}}B$, and the preceding $k-1$
edges form a path in $G[B]$. Hence
$$
d_{G[B]}(u,\partial_{\mathrm{in}}B)\leq k-1,
$$
which proves the reverse inequality. The graph is
connected and $B^c$ is nonempty, so these distances
are finite. The representation follows by comparing
both sides with $r$.
\end{proof}

\begin{proof}[Proof of Theorem~\ref{thm:graph_boundary}]
By the definition of $\Gamma_B(r-1)$, every nonempty
$S\subseteq B$ satisfies
$$
|\{u\in B:d_{G[B]}(u,S)\leq r-1\}|
\leq
\Gamma_B(r-1)|S|.
$$
Since $G$ is connected and $B$ is nonempty and proper,
$\partial_{\mathrm{in}}B$ is nonempty. Taking
$S=\partial_{\mathrm{in}}B$ and using
\eqref{eq:graph_boundary_representation} gives
$$
\begin{aligned}
|\operatorname{bd}_G(B;r)|
&=
|\{u\in B:
d_{G[B]}(u,\partial_{\mathrm{in}}B)\leq r-1\}|\\
&\leq
\Gamma_B(r-1)|\partial_{\mathrm{in}}B|.
\end{aligned}
$$
Every vertex in $\partial_{\mathrm{in}}B$ is incident to
at least one edge in $E(B,B^c)$. Counting crossing edges
by their endpoint in $B$ therefore
$
|\partial_{\mathrm{in}}B|\leq|E(B,B^c)|.
$
Moreover, $\operatorname{bd}_G(B;r)\subseteq B$, so
$|\operatorname{bd}_G(B;r)|\leq |B|$. Together with
the preceding bounds, this gives
$$
\begin{aligned}
|\operatorname{bd}_G(B;r)|
&\leq \Gamma_B(r-1)|\partial_{\mathrm{in}}B|\\
&\leq \Gamma_B(r-1)|E(B,B^c)|.
\end{aligned}
$$
And then
$$
|\operatorname{bd}_G(B;r)|
\leq
\min\left\{
|B|,\,
\Gamma_B(r-1)|E(B,B^c)|
\right\}.
$$
Dividing by $|B|$ yields
$$
\begin{aligned}
\frac{|\operatorname{bd}_G(B;r)|}{|B|}
&\leq
\min\left\{
1,\,
\frac{\Gamma_B(r-1)|E(B,B^c)|}{|B|}
\right\}
% \\
% &=
% \min\left\{
% 1,\,
% \frac{\Gamma_B(r-1)\Phi_G(B)
% \min\{\operatorname{vol}_G(B),
%        \operatorname{vol}_G(B^c)\}}{|B|}
% \right\},
\end{aligned}
$$
which complete the proof.
% where the final equality follows from the definition of
% the conductance $\Phi_G(B)$.
\end{proof}

To obtain a bound without the growth function $\Gamma_B$,
let
$
\deg_{\max}(B)
:=
\max_{u\in B}\deg_{G[B]}(u)
$
be the maximum internal degree, where
$\deg_{G[B]}(u):=|\{v\in B:\{u,v\}\in E\}|$.
For each integer $t\geq0$,
define
$$
M_B(t)
:=
\begin{cases}
1,
& \deg_{\max}(B)=0,\\[2mm]
1+\deg_{\max}(B)
\displaystyle\sum_{j=0}^{t-1}
\bigl(\deg_{\max}(B)-1\bigr)^j,
& \deg_{\max}(B)\geq1,
\end{cases}
$$
where the sum is empty when $t=0$ and $0^0=1$.

\begin{lem}[Growth under bounded internal degree]
\label{lem:graph_degree_growth}
For every nonempty $S\subseteq B$ and integer $t\geq0$,
$$
|\{u\in B:d_{G[B]}(u,S)\leq t\}|
\leq M_B(t)|S|.
$$
In particular, $\Gamma_B(t)\leq M_B(t)$.
\end{lem}

\begin{proof}
If $\deg_{\max}(B)=0$, every vertex of $G[B]$ is
isolated. The $t$-neighborhood of $S$ is then $S$
itself, and the result follows from $M_B(t)=1$. Suppose $\deg_{\max}(B)\geq1$. For $j\geq0$, let
$$
L_j:=\{u\in B:d_{G[B]}(u,S)=j\}.
$$
In particular, $L_0=S$. Every vertex in $L_1$ has
an edge to $L_0$, while each vertex of $L_0$ has at
most $\deg_{\max}(B)$ neighbors. Hence
$
|L_1|
\leq \deg_{\max}(B)|L_0|
=\deg_{\max}(B)|S|.
$
Now for $j\geq1$ we derive two bounds based on the same count
$|E(L_j,L_{j+1})|$, i.e. the number of edges crossing
$L_j$ to $L_{j+1}$. 

First, every vertex in $L_{j+1}$ has
at least one neighbor in $L_j$, so
$
|L_{j+1}|\leq |E(L_j,L_{j+1})|.
$
Next, for each vertex in $L_j$, deducting at least one
neighbor which lies in $L_{j-1}$ along a shortest path
to $S$, it has at most $\deg_{\max}(B)-1$ neighbors in
$L_{j+1}$ which gives
$
|E(L_j,L_{j+1})|
\leq(\deg_{\max}(B)-1)|L_j|.
$
Combining these inequalities yields
$$
|L_{j+1}|\leq(\deg_{\max}(B)-1)|L_j|.
$$
Starting from the bound on $|L_1|$ and applying this
recurrence $j-1$ times, we obtain, for every $j\geq1$,
$$
|L_j|
\leq
(\deg_{\max}(B)-1)^{j-1}|L_1|
\leq
\deg_{\max}(B)(\deg_{\max}(B)-1)^{j-1}|S|.
$$

The layers $L_0,\ldots,L_t$ are disjoint, and their
union is the set of vertices within distance $t$ of
$S$ in $G[B]$. Therefore
$$
\begin{aligned}
|\{u\in B:d_{G[B]}(u,S)\leq t\}|
&=\sum_{j=0}^{t}|L_j|\\
&=|S|+\sum_{j=1}^{t}|L_j|\\
&\leq
|S|+
\sum_{j=1}^{t}
\deg_{\max}(B)
\bigl(\deg_{\max}(B)-1\bigr)^{j-1}|S|\\
&=
|S|\left(
1+\deg_{\max}(B)
\sum_{i=0}^{t-1}
\bigl(\deg_{\max}(B)-1\bigr)^i
\right)\\
&=M_B(t)|S|.
\end{aligned}
$$

Finally, since the bound holds for every nonempty
$S\subseteq B$,
$$
\begin{aligned}
\Gamma_B(t)
&=
\max_{\varnothing\neq S\subseteq B}
\frac{|\{u\in B:d_{G[B]}(u,S)\leq t\}|}{|S|}\\
&\leq
\max_{\varnothing\neq S\subseteq B}M_B(t)
=M_B(t).
\end{aligned}
$$
\end{proof}

Alternatively to connect explicitly with maximum-degree, we write $\operatorname{vol}_G(S):=\sum_{u\in S}\deg_G(u)$.
The conductance of $B$ in $G$ is
$$
\Phi_G(B):=
\frac{|E(B,B^c)|}
{\min\{\operatorname{vol}_G(B),
       \operatorname{vol}_G(B^c)\}}.
$$
\begin{cor}[Explicit maximum-degree bound]
\label{cor:graph_degree_boundary}
For every integer $r\geq1$,
$$
\frac{|\operatorname{bd}_G(B;r)|}{|B|}
\leq
\min\left\{
1,\,
\frac{M_B(r-1)|E(B,B^c)|}{|B|}
\right\}
\leq
\min\left\{
1,\,
M_B(r-1)\Phi_G(B)\overline d_G(B)
\right\},
$$
where $\overline d_G(B)=\operatorname{vol}_G(B)/|B|$.
\end{cor}

\begin{proof}
Applying Lemma~\ref{lem:graph_degree_growth} in
Theorem~\ref{thm:graph_boundary} proves the
first inequality in the corollary. For the second inequality, the definition of
$\Phi_G(B)$ gives
$$
\begin{aligned}
|E(B,B^c)|
&=
\Phi_G(B)
\min\{\operatorname{vol}_G(B),
       \operatorname{vol}_G(B^c)\}\\
&\leq
\Phi_G(B)\operatorname{vol}_G(B)\\
&=
\Phi_G(B)|B|\overline d_G(B).
\end{aligned}
$$
Consequently,
$$
\frac{M_B(r-1)|E(B,B^c)|}{|B|}
\leq
M_B(r-1)\Phi_G(B)\overline d_G(B).
$$
Taking the minimum of each side with $1$ proves
the second inequality.
\end{proof}

\subsection{Additional Related Work}
\label{app:related_work}

\paragraph{Conformal Prediction (CP), Weighted CP (WCP), and Localized CP (LCP)}
Unlike single-point predictors that do not quantify uncertainty \cite{shafer2008tutorial}, CP provides a simple distribution-free framework for constructing prediction sets in classification and prediction intervals in regression with finite-sample guarantees \cite{shafer2008tutorial, angelopoulos2021gentle, maincp2005}. CP's validity comes from the assumption of exchangeability, and as a distribution-free method, it does not rely on strong parametric assumptions or strict distributions. \cite{ranklist2021, maincp2005}. The high variance in marginal coverage across different regions of the data distribution has increased interest in conditional CP although exact conditional coverage cannot generally be achieved in finite samples \cite{infeasiblecondcp2020}. 

Many recent works study approximate or relaxed conditional coverage to improve coverage in heterogeneous settings by adapting prediction set sizes to the difficulty of individual instances \cite{validconfidenseset2021, validadaptiveacoverage2020}. Also, several weighted CP methods have proposed relaxing standard exchangeability assumptions \cite{wcp2019, beyondexchangeability2022}. These approaches typically estimate distribution ratios between calibration and test samples and use them to compute weighted quantiles \cite{rlcp2024, callcp2022}. 

Similarly, localized CP (LCP) gives larger weights to calibration points that are more similar to the test instance to improve local coverage behavior \cite{wcp2019, rlcp2024}. From this category, Randomized LCP (RLCP) further introduces randomization and anchor sampling strategies to improve coverage around the test point itself rather than only its surrounding neighborhood \cite{rlcp2024}. Still, most localized methods depend heavily on the quality of the feature representation used to measure similarity \cite{highdimensionalcovarite2025, rlcp2024}. In high-dimensional settings or when embeddings are poorly structured, defining reliable local neighborhoods becomes difficult and can lead to inefficient prediction intervals \cite{beyondexchangeability2022, highdimensionalcovarite2025}.

\paragraph{Uncertainty Quantification in Graph Neural Networks (GNNs)}
Calibration remains a challenge in graph neural networks (GNNs) since 1) their learned representations are often not sufficiently separable or reliable \cite{cagcn2021, gats2022}, and 2) predictions are not uniformly reliable across different regions of the graph due to structural effects \cite{naps2023, gats2022}. This is particularly important in high-stakes applications such as drug discovery or fraud detection, where inaccurate confidence estimates may lead to unreliable decisions \cite{feinberg2018potential, cheng2025graph, gats2022}. In addition, under heterophily, GNNs can suffer from over-squashing, while deeper architectures are also prone to over-smoothing \cite{oversmoothingoversquashingtradeoff2023, oversmoothingsurvey2023}. As a result, node embeddings may become noisy or less informative for measuring similarity, which limits the application of existing LCP methods to graph data as they rely heavily on stable embedding-space distances \cite{oversmoothingoversquashingtradeoff2023, rlcp2024}.

\paragraph{Conformal Prediction on Graphs}
Applying CP to graph data is challenging because node dependencies violate the standard i.i.d.\ assumption \cite{networkcp2023, naps2023}. Still, prior work has shown that exchangeability can hold in transductive settings for conventional GNNs due to permutation invariance, which supports the use of finite-sample CP guarantees on graph data \cite{ranklist2021, cfgnn2023, networkcp2023}. This has motivated CP for graph tasks such as link prediction and out-of-distribution detection \cite{oodcpgnn2025, linkfdr2023, cplinprediction2024}.

Recent graph CP methods often incorporate additional structural or contextual information during calibration \cite{naps2023, cfgnn2023, rrgnn2025}. Some approaches use embedding-space density estimates for weighting \cite{codrug2023}, while others rely on structural partitioning \cite{rrgnn2025}, graph diffusion scores \cite{cpforgnn2023}, or combinations of node features and graph topology to adapt nonconformity scores \cite{snaps2024}. Many of these methods remain sensitive to embedding quality or rigid graph partitions. Our approach instead uses a structure-aware kernel to improve localization when feature-space similarity is unreliable.

\subsection{Limitations of RLCP}\label{app:limitations}

While RLCP excels in low-dimensional settings, its application to GNNs is hindered by poor embedding discriminability. GNN embeddings often exhibit high feature correlation, non-uniformity, and over-smoothing. Consequently, applying a Gaussian kernel here tends to collapse the distribution toward a Dirac delta or a uniform weight. We verify this behavior by measuring the weights of the calibration and test samples across varying bandwidths.

\begin{figure}[ht]
    \centering
    \includegraphics[width=.7\linewidth]{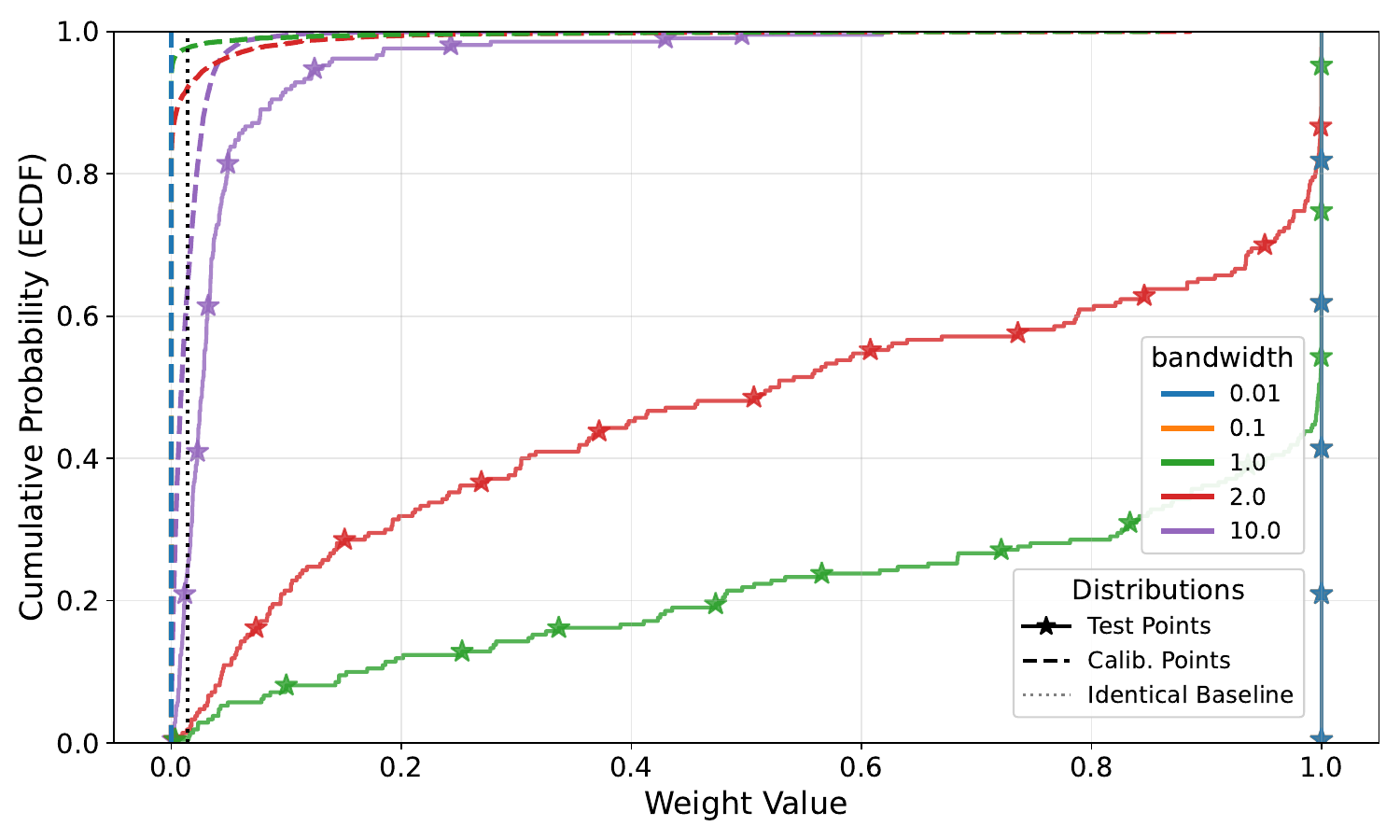}

    \vspace{-2pt}
    \caption{
    GOODCBAS dataset: Impact of Gaussian kernel bandwidth $h$ on RLCP weight distribution. Low bandwidths concentrate probability mass on test points, leaving calibration points with near-zero weights. Conversely, large bandwidths cause weights to jump toward a uniform distribution where all points receive near-zero values.}
    \label{fig:rlcpweight}
\end{figure}

Figure \ref{fig:rlcpweight} shows the empirical cumulative distribution function (ECDF) for calibration (dashed) and test (solid) weights according to sampled anchor points for a Gaussian kernel applied in the embedding space for different bandwidths. Small bandwidth assigns almost all the weight to the test points (i.e., $w_{n+1} \simeq 1.0$), leaving the calibration nodes with identical weights close to zero, i.e., $w_i \simeq 0.0: \forall i\in [n]$. This pushes the estimated quantile $\hat{q}_{1-\alpha}(\xv_{n+1}, \tilde{\xv}_{n+1}) \simeq +\infty$ to infinity according to Eq. \ref{eq:rlcpquantile}. Whereas, larger bandwidths yield identical weights for all points due to indiscriminative, over-smoothed embeddings. This reduces RLCP to standard SCP, ignoring local proximity for better test conditional coverage. We address this by introducing \ourmethod, which explicitly incorporates graph-structural proximity.

\subsection{Empirical Insights and Design Principle}\label{app:homophily}
\begin{figure} 
    \centering
    \includegraphics[width=.8\linewidth]{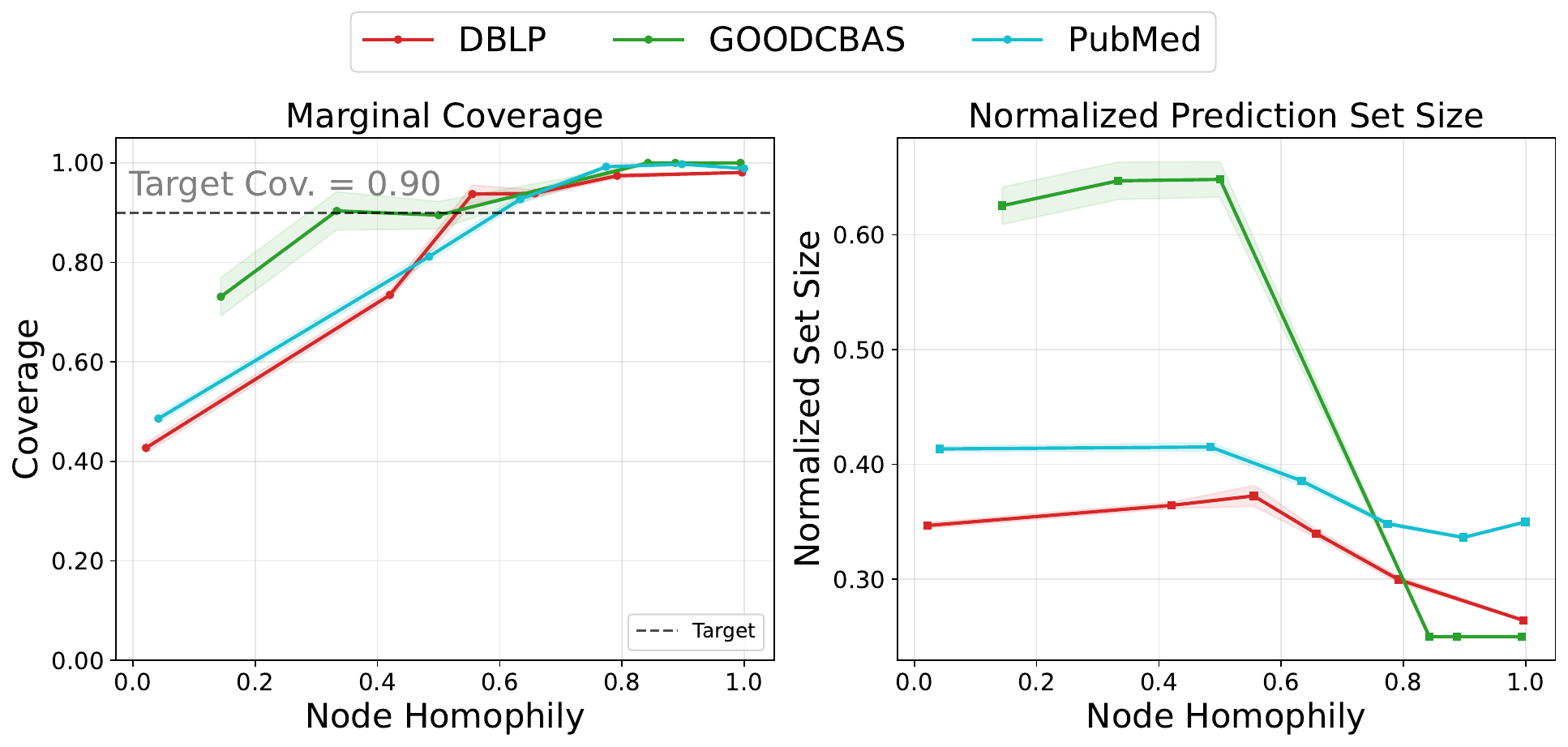}

    \vspace{-2pt}
    \caption{
    Impact of node homophily on the marginal coverage and prediction length/size of SCP. Satisfying coverage for more homophilous nodes is easier at the same or lower prediction length, underlining the significant contribution of graph structure in CP application for GNNs.
    }
    \label{fig:homophily_coverage_pi_classification}
\end{figure}

Our assumption is that weighted CP can highly benefit from nodes' structural dependencies. We confirm this assumption by an experiment that shows the effect of node homophily on the coverage and prediction length of vanilla SCP (defined in Sec. \ref{sec:CP}). For classification datasets, a node's homophily ratio $\eta_u=\frac{|\{v \in \nb_u: y_v=y_u\}|}{|\nb_u|}$ and for regression datasets, we have $\eta_u = 1 - \frac{1}{|\mathcal{N}_u|} \sum_{v \in \mathcal{N}_u} \frac{|y_v - y_u|}{\underset{i,j:A_{ij}=1}{\max} |y_i - y_j|}$, where $y_u$ is target value (e.g. label) of $u$. Figure \ref{fig:homophily_coverage_pi_classification} elaborates on a common pattern: nodes with low homophily suffer from significant under-coverage, while high-homophily nodes are well covered, exhibiting an easier task. Also, the prediction lengths are generally stable or decrease with higher homophily ratios. This demonstrates that SCP fails to guarantee coverage for heterophilous nodes, even with a higher uncertainty budget. 

Because uncertainty and graph topology are strongly correlated, we explicitly incorporate the underlying structure into our quantile weighting. \ourmethod replaces embedding-based kernels with a structure-aware kernel that governs both anchor sampling and weighting. By defining proximity through topology, we enable localized CP that respects structural dependencies and bypasses the suboptimality of message-passing embeddings \cite{skipconnection2023}. \ourmethod comprises three components: (1) a Personalized PageRank-based localization kernel, (2) randomized anchor sampling and weighting compatible with RLCP, and (3) a feature-aware densification step to mitigate sparsity. These are detailed below.

%\begin{wrapfigure}[17]{r}{0.5\textwidth} 
% The [12] tells LaTeX to wrap exactly 12 lines of text
% {r} for right side, {l} for left

\subsection{Pseudocode for Our Method}
\label{app:pseudocode}
Algorithm \ref{alg:ourmethod} presents the Pseudocode of the three main stages of our proposed method \ourmethod: 1) \textit{graph densification} with a PCA projection and adaptive bandwidths, 2) \textit{anchor point sampling} according to a geometric distribution of random walk lengths, and 3) \textit{weighting and quantile computation} using our structure-aware kernel. The algorithm shows the PCA-enabled variant; otherwise, Gaussian similarities are computed directly on the original embeddings.

\begin{algorithm}[!ht]
   \caption{\ourmethod}
   \label{alg:ourmethod}
\begin{algorithmic}[1]
   \STATE {\bfseries Input:} Graph $G=(V,E)$ with adjacency $\mathbf{A}$ and features $\mathbf{X}$, frozen GNN embeddings, Calibration $\mathcal{D}_{cal}$, test nodes $V_{\mathrm{test}}$ (including $u_{n+1}$), Params $\alpha, \beta, K, c, h$, fixed $\tau_d$
   \STATE {\bfseries Output:} Prediction set $\widehat{C}_n(\xv_{n+1})$
   \vspace{2mm}
   \STATE \textbf{Stage 1: Feature-Aware Densification}
   \STATE $V_{\mathrm{pool}}\leftarrow V_{\mathrm{cal}}\cup V_{\mathrm{test}}$
   \STATE $\mathbf{V}_{c}, \mathbf{\Lambda}_{c}, \bar{\xv}_{\mathrm{pool}} \leftarrow \text{PCA}(\{\xv_i:i\in V_{\mathrm{pool}}\}, \text{top-}c)$
   \STATE $\tilde{\mathbf A}\leftarrow\mathbf A$
   \FOR{each pair $(i, j)$ with $i\neq j$ and $\mathbf{A}_{ij}=0$}
        \STATE $\mathbf{h}_i = \mathbf{V}_{c}^T(\xv_i-\bar{\xv}_{\mathrm{pool}})$;  $\mathbf{h}_j = \mathbf{V}_{c}^T(\xv_j-\bar{\xv}_{\mathrm{pool}})$
       \STATE $s_{ij} \leftarrow \exp\left(-\frac{1}{2}(\mathbf{h}_i-\mathbf{h}_j)^T (h^2 \mathbf{\Lambda}_c)^{-1} (\mathbf{h}_i-\mathbf{h}_j)\right)$
       \IF{$s_{ij} \geq \tau_d$}
           \STATE $\tilde{\mathbf{A}}_{ij} \leftarrow s_{ij}$
       \ENDIF
   \ENDFOR
   \vspace{2mm}
   \STATE \textbf{Stage 2: Anchor Sampling}
   \STATE For any $i$ with $\sum_j\tilde{\mathbf A}_{ij}=0$, set $\tilde{\mathbf A}_{ii}\leftarrow1$
   \STATE $\tilde{\mathbf D}_{ii}\leftarrow\sum_j\tilde{\mathbf A}_{ij}$; $\mathbf{M} \leftarrow \tilde{\mathbf{D}}^{-1}\tilde{\mathbf{A}}$
   \STATE Draw $L\sim\operatorname{Geom}(\beta)$ with $\mathbb P(L=k)=\beta(1-\beta)^k$, $k=0,1,\ldots$
   \STATE Set $L_K\leftarrow\min(L,K)$ and $X_0\leftarrow u_{n+1}$
   \FOR{$t=0,\ldots,L_K-1$}
       \STATE Sample $X_{t+1}\sim\mathbf M_{X_t,\cdot}$
   \ENDFOR
   \STATE $\tilde{u}_{n+1}\leftarrow X_{L_K}$
   \vspace{2mm}
   \STATE \textbf{Stage 3: Weighting \& Quantile Computation}
   \STATE Initialize $\boldsymbol\pi^{(0)}\leftarrow\mathbf e_{\tilde u_{n+1}}$
   \FOR{$t=0,\ldots,K-1$}
       \STATE $\boldsymbol\pi^{(t+1)}\leftarrow(1-\beta)\boldsymbol\pi^{(t)}\mathbf M+\beta\mathbf e_{\tilde u_{n+1}}$
   \ENDFOR
   \STATE $\tilde{\boldsymbol{\pi}}_{n+1}\leftarrow\boldsymbol\pi^{(K)}$
   \FOR{$i \in \mathcal{D}_{cal} \cup \{u_{n+1}\}$}
       \STATE $\tilde{w}_i \leftarrow [\tilde{\boldsymbol{\pi}}_{n+1}]_i / d_i$ \quad \COMMENT{$d_i=\tilde{\mathbf D}_{ii}$}
   \ENDFOR
   \STATE Normalize weights $\tilde{w}_i \leftarrow \tilde{w}_i / \sum \tilde{w}_j$
   \STATE $\hat{q} \leftarrow \text{Quantile}_{1-\alpha}\left(\sum \tilde{w}_i \delta_{s(\xv_i, y_i)} + \tilde{w}_{n+1}\delta_{\infty}\right)$
   \STATE \textbf{return} $\{y \in \mathcal{Y} \mid s(\xv_{n+1}, y) \leq \hat{q}\}$
\end{algorithmic}
\end{algorithm}

\subsection{Fixed Densification Threshold.}\label{app:fixed_densification_threshold}
The similarity threshold $\tau_d$ is fixed before calibration and evaluation and is not selected using calibration or test labels. When enabled, PCA is fitted using the pooled unlabeled calibration and test embeddings, and the same transformation is applied to all nodes. The threshold and transformation are therefore unchanged by swapping calibration/test roles within this pooled set.

\subsection{Experimental Settings}\label{app:experiments}
For all datasets, we split each dataset into 4 sets: training, validation, calibration, and test. We train the models on the training set and use the validation set to tune the hyperparameters. Once the models are trained, we fix the pretrained models. The selection fractions (train, validation, calibration, test) are as follows: ANH, CHI, TWT: (0.4, 0.1, 0.1, 0.4) - CMP, CRA, DBLP, PMD, : (0.2, 0.1, 0.2, 0.5) - EDU, ELC, INC, UNM: (0.4, 0.1, 0.1, 0.4) - CBAS, WKB: (0.5, 0.1, 0.1, 0.3). We use GCN \cite{gcn2017}, as our pretrained model. Across all the experiments, we fix: miscoverage $\alpha=0.1$, the number os APPNP steps, $K=30$, the base badnwith $h=2.0$, and $c=5$ wherever PCA is used. We choose $\tau_d \in \{0.2, 0.5, 0.8\}$, $\beta \in \{0.2, 0.4\}$. We run the experiments on a single GPU core NVIDIA GeForce RTX 3090 and x86\_64 CPUs. 

\paragraph{Design Principle of PCA Application} \label{app:pca_application}
PCA is enabled on eight datasets: all seven regression datasets and one classification dataset (WKB). We apply PCA when reducing redundancy can improve similarity estimation in high dimensional embeddings. However, when the original graph provides sufficient local connectivity, PPR can capture informative structural dependencies without aggressive densification. In such cases, dimensionality reduction may increase the apparent similarity of distinct nodes, potentially introducing noisy edges. We therefore retain the original embeddings to favor more conservative densification.
% The code is available here: \url{https://anonymous.4open.science/r/GraphLCP-D208}.

\begin{figure}[t]
    \centering
    \captionsetup[subfigure]{font=small, skip=3pt}

    \begin{subfigure}[b]{0.45\textwidth}
        \centering 
        {\includegraphics[width=\textwidth, trim={0 0 0 0}, clip]{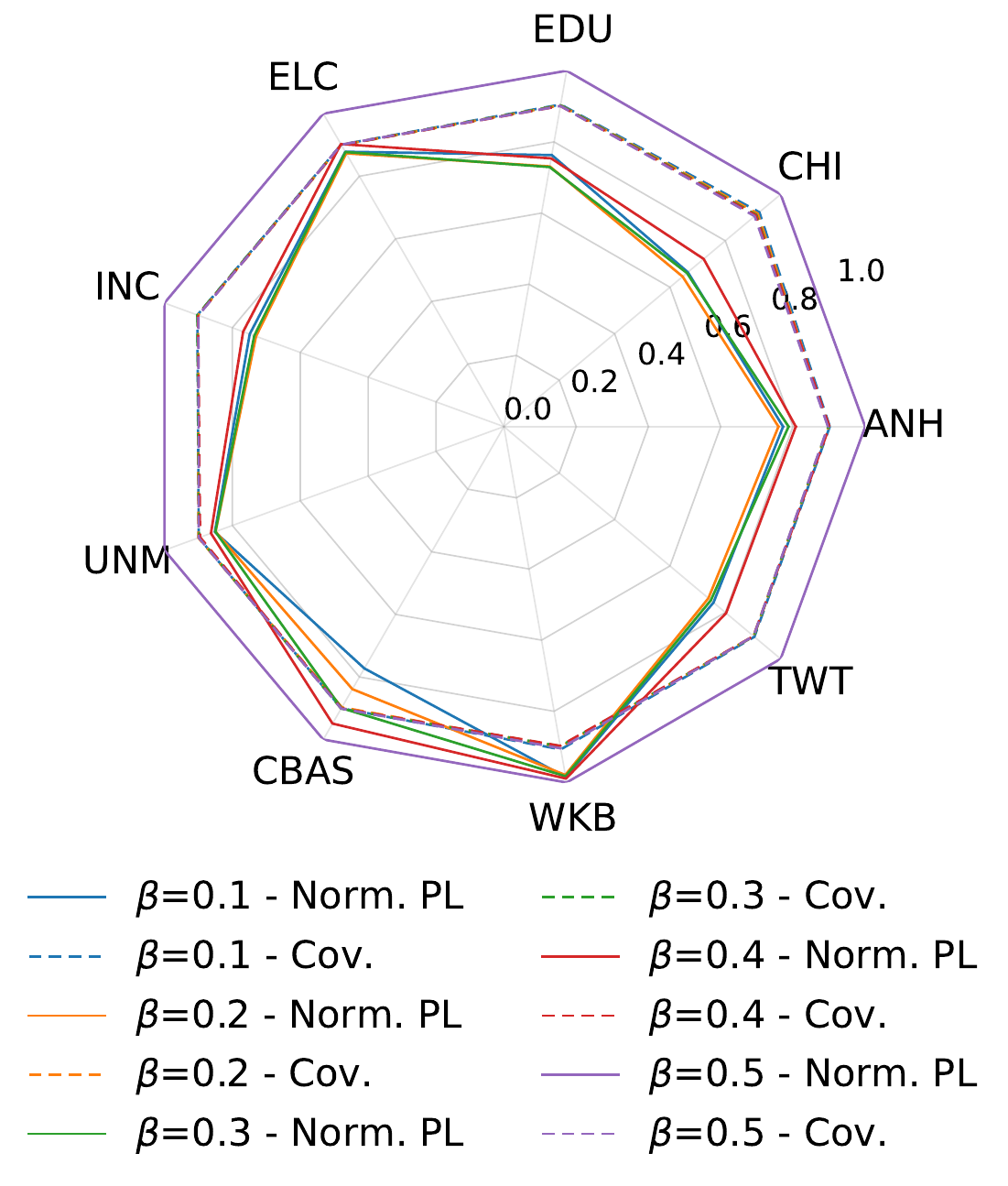}}
        \caption{Marginal Cov. and PL}
    \end{subfigure}
    \hfill
    \begin{subfigure}[b]{0.45\textwidth}
        \centering 
        \includegraphics[width=\textwidth, trim={0 0 0 0}, clip]{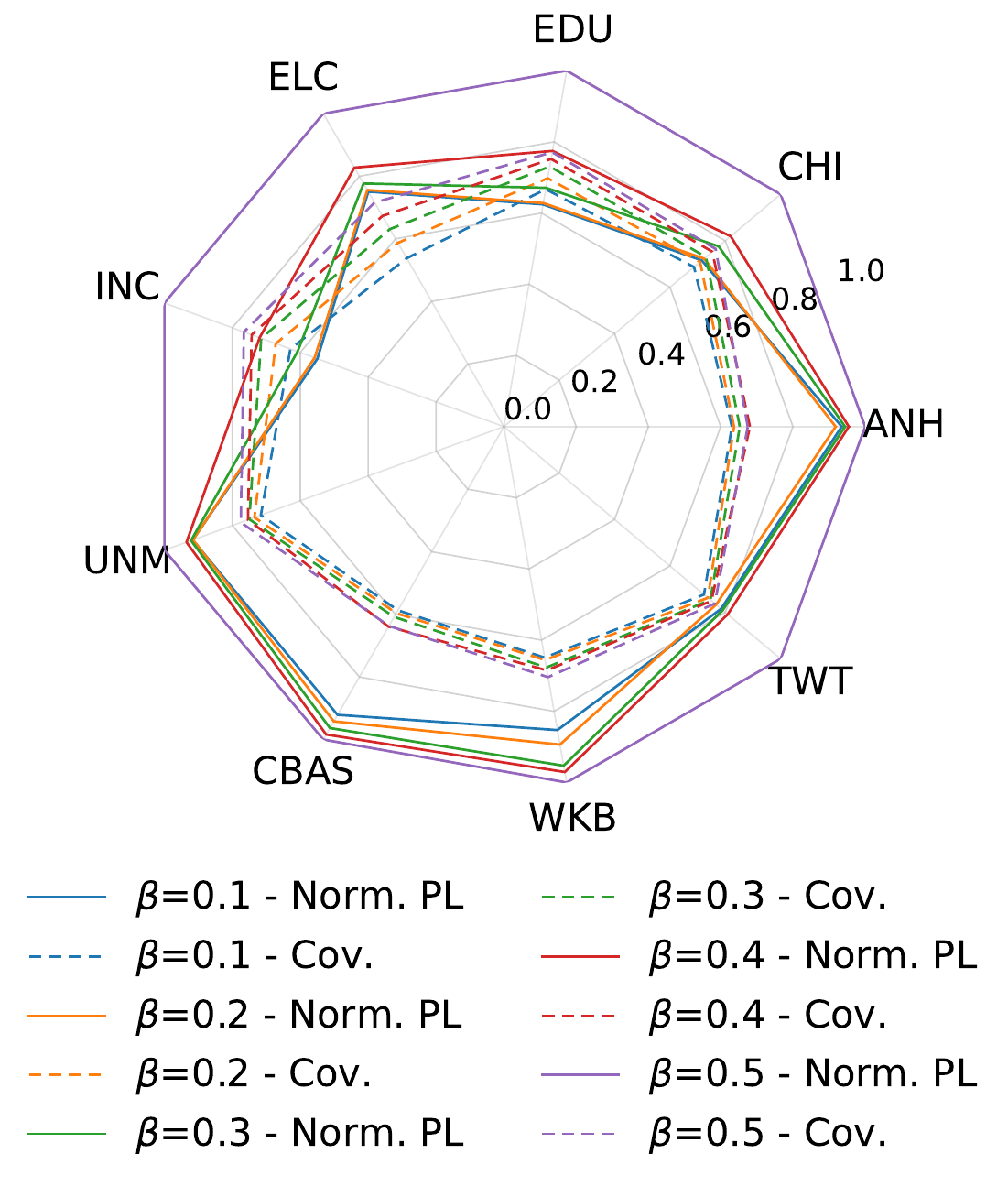}
        \caption{WSC Cov. and PL}
    \end{subfigure}

    \vspace{-2pt}
    \caption{Sensitivity analysis of the PPR restart probability $\beta$. Higher $\beta$ values prioritize local neighborhood information, leading to improved marginal and WSC coverage at the cost of a moderate increase in WSC PL.}
    \label{fig:ppr_alpha}
\end{figure}

\subsection{Additional Ablation Studies}

The previous ablation studies clarified the significant impact of the PPR kernel and the densification step. Therefore, here analyze the effect of the two main hyperparameters within the setup of the PPR kernel and the densification. 

\paragraph{PPR restart probability ($\beta$)} First, we conduct a sensitivity analysis over the PPR restart probability $\beta\in [0.1, 0.5]$. This hyperparameter controls the kernel's locality extent: lower values allow longer random walks, allowing more nodes to contribute to the weighted CP and vice versa. The results in Figure \ref{fig:ppr_alpha} show that increasing $\beta$ improves both marginal and WSC (coverage) while moderately increasing PL (prediction length) of the worst-slab scenario. In other words, smaller $\beta$ yields overly global kernels, whereas larger values emphasize locality. Overall, performance is stable across this range, with intermediate values providing the best balance.

\begin{figure}[t]
    \centering
    \captionsetup[subfigure]{font=small, skip=3pt}

    \begin{subfigure}[b]{0.375\textwidth}
        \centering
        {\includegraphics[width=\textwidth, trim={0 0 0 0}, clip]{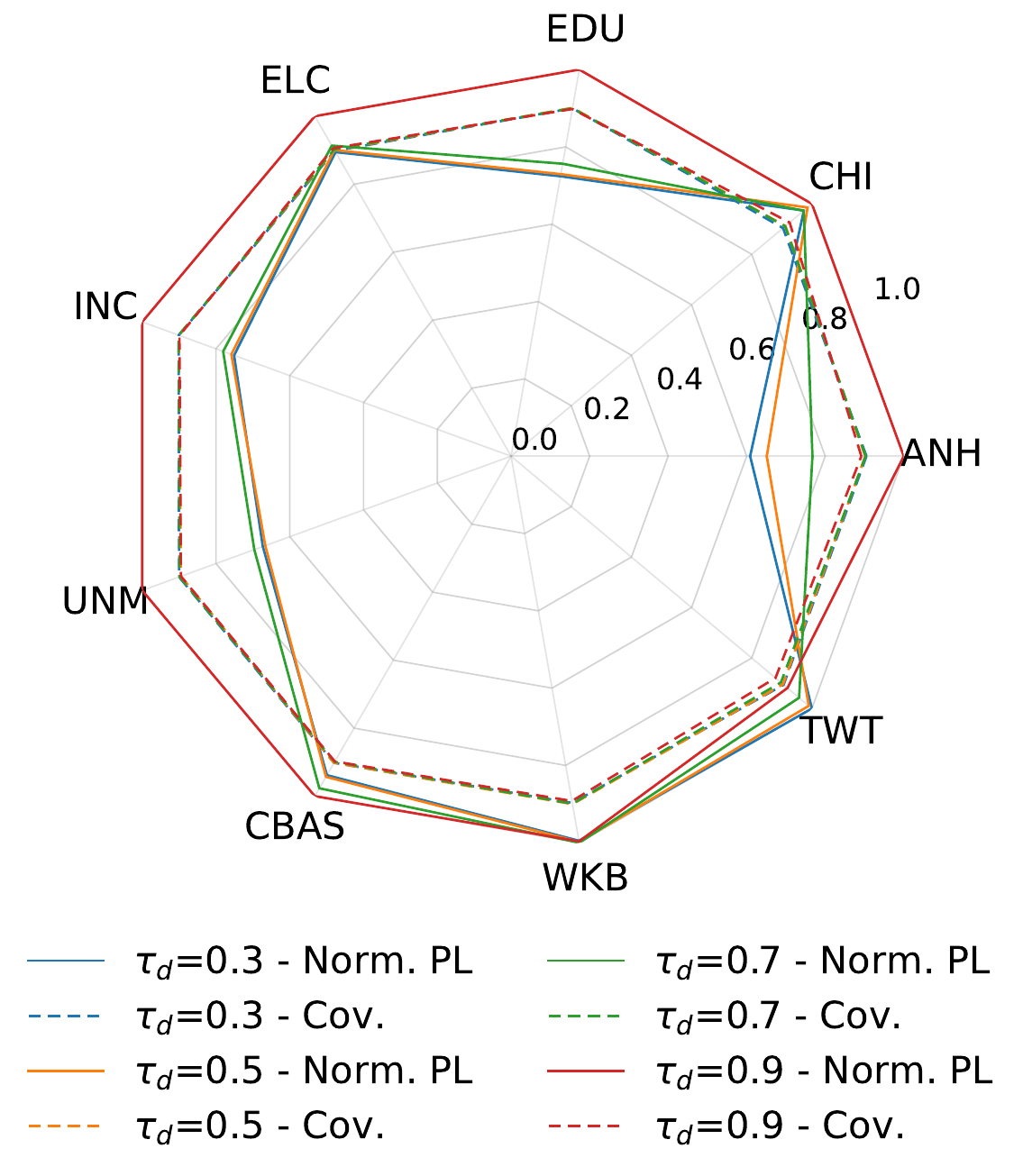}}
        \caption{Marginal Cov. and PL}
    \end{subfigure}
    \hfill
    \begin{subfigure}[b]{0.375\textwidth}
        \centering
        \includegraphics[width=\textwidth, trim={0 0 0 0}, clip]{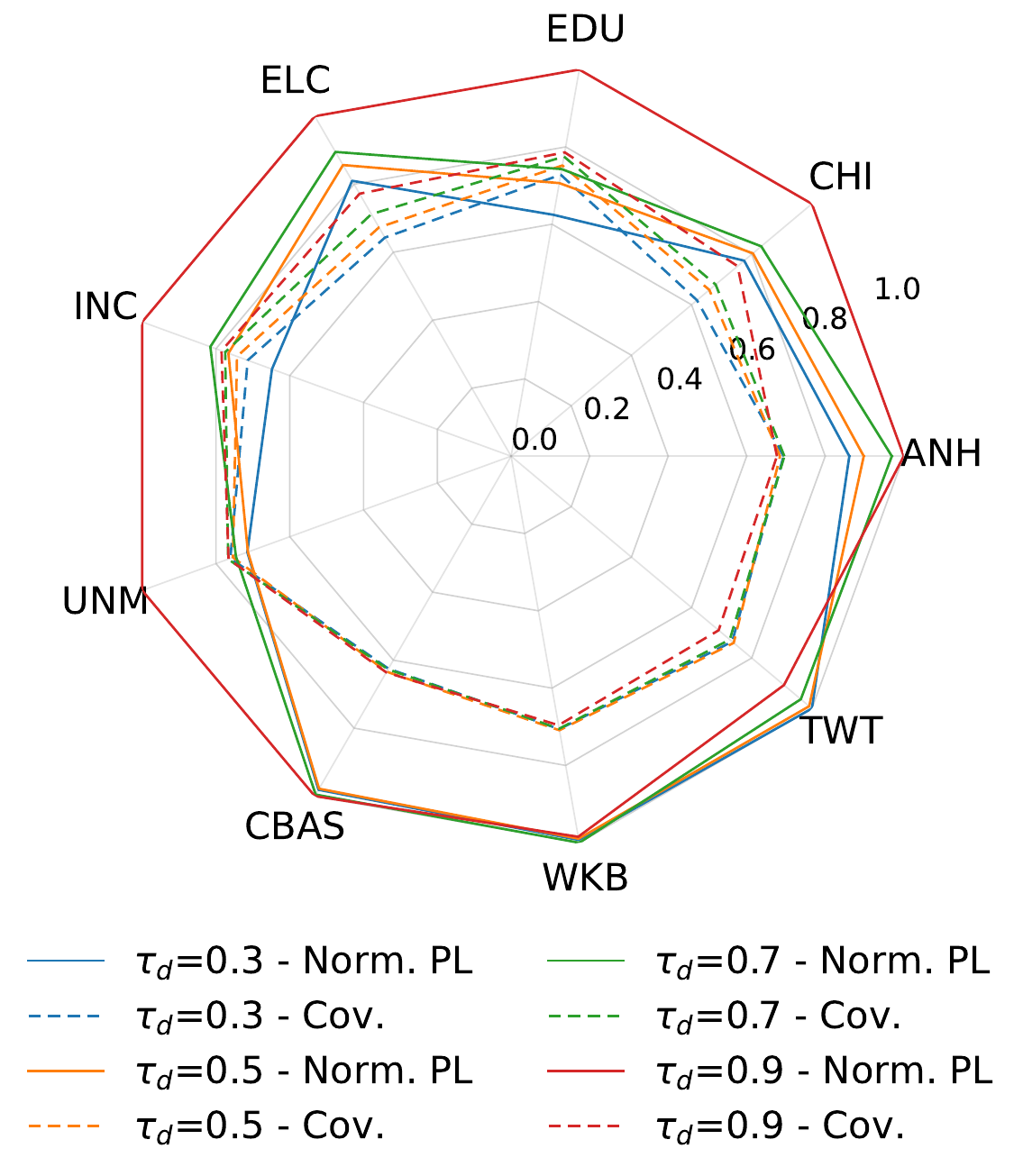}
        \caption{WSC Cov. and PL}
    \end{subfigure}
    \hfill
    \begin{subfigure}[b]{0.125\textwidth}
        \centering
        \raisebox{50pt}{\includegraphics[width=\textwidth, trim={0 0 0 0}, clip]{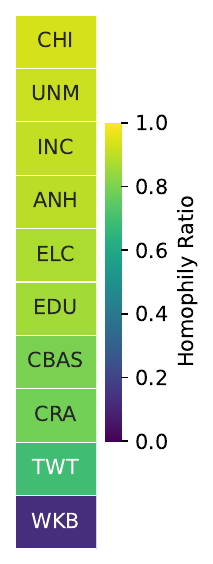}}
    \end{subfigure}

    \vspace{-2pt}
    \caption{Sensitivity analysis of the fixed densification threshold $\tau_d$. Coverage and prediction length vary with the extent of similarity-based edge additions and graph homophily.}
    \label{fig:dens_thresh}
\end{figure}

\paragraph{Fixed densification threshold ($\tau_d$)}
We examine the sensitivity to the fixed threshold $\tau_d\in\{0.3,0.5,0.7,0.9\}$ across 10 datasets. Lower thresholds add more similarity-based edges, while higher thresholds restrict additions to more similar nodes, reducing potentially noisy connections. Figure~\ref{fig:dens_thresh} shows that the effect varies with graph homophily: lower thresholds favor coverage on less homophilous graphs (e.g., WKB), whereas higher thresholds can improve coverage on more homophilous graphs (e.g., ELC). Overall, $\tau_d=0.5$ provides a favorable balance between coverage and prediction length. Homophily is used only to interpret these results, not to select the threshold.

\begin{figure}[t]
    \centering
    \includegraphics[width=0.4\linewidth]{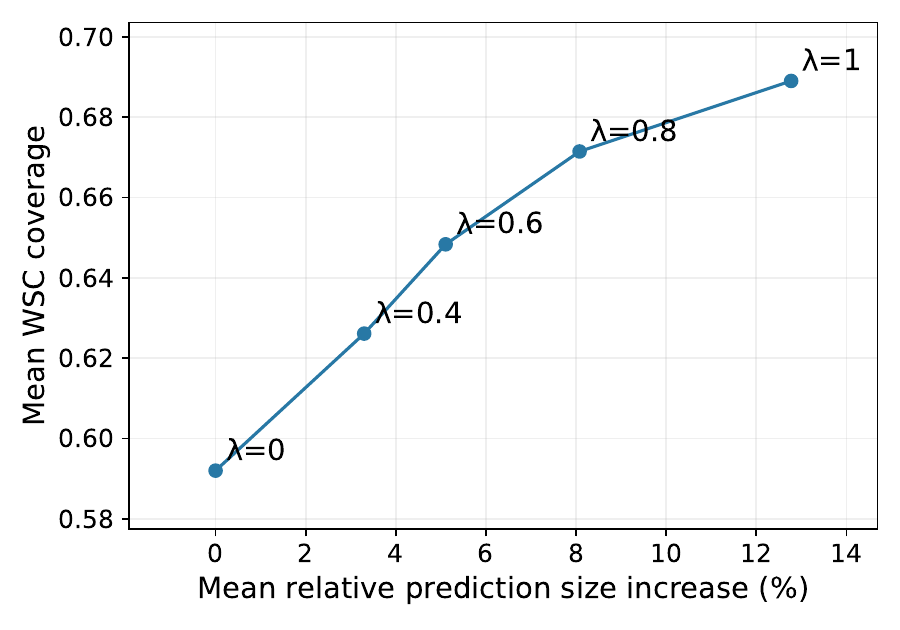}
    \caption{WSC--prediction-size trade-off under
    mixed uniform--PPR localization across nine datasets.
    Prediction size is normalized by its value at
    $\lambda=0$ within each dataset before averaging.}
    \label{fig:localization_tradeoff}
\end{figure}

\paragraph{WSC--prediction-size trade-off.}
\label{app:localization_tradeoff}
To examine the trade-off between structural localization
and prediction size, we interpolate between uniform
and PPR kernels:
\begin{equation}
    H_\lambda(u,v)
    =(1-\lambda)\frac{1}{N}
    +\lambda H_K(u,v),
    \qquad \lambda\in[0,1].
\end{equation}
We use the same mixed kernel for anchor sampling and
calibration weighting. Here, $\lambda=0$ recovers SCP
and $\lambda=1$ recovers \ourmethod, while intermediate
values serve only as a diagnostic. Figure~\ref{fig:localization_tradeoff}
shows that stronger structural localization consistently
improves mean WSC, with a gradual increase in prediction
size across nine datasets (ANH, CHI, EDU, ELC, INC, UNM, TWT, CBAS, WKB). This highlights the coverage--efficiency trade-off, where the preference for
stronger localization depends on the cost of local
undercoverage.

\subsection{More Conditional and Marginal Results}\label{app:addiotnalexperiments}

\paragraph{WSC results}
To have a more detailed picture of how we perform in comparison to the baselines, we show the full results of WSC in Tables \ref{tab:wsc_coverage_all} and \ref{tab:wsc_pred_length_all}. As previously discussed, \ourmethod attains a desirable trade-off between prediction length and coverage in comparison with other baselines, and this trade-off is clearer on the regression tasks. 

\begin{table*}[t]
\scriptsize
\centering
\caption{WSC results. \ourmethod achieves strong WSC across regression and classification datasets. ``-'' means infinite-length prediction interval. OOT (out-of-time) shows results attained after an unreasonably long time.}
\vspace{-1mm}
\resizebox{\textwidth}{!}{
    \begin{tabular}{L{0.6in}C{0.3in}C{0.3in}C{0.3in}C{0.3in}C{0.3in}C{0.3in}C{0.3in}C{0.3in}C{0.3in}C{0.3in}C{0.3in}C{0.3in}C{0.3in}}
        \toprule
        \textbf{Method} & \textbf{ANH} & \textbf{CHI} & \textbf{EDU} & \textbf{ELC} & \textbf{INC} & \textbf{UNM} & \textbf{TWT} & \textbf{CMP} & \textbf{CRA} & \textbf{DBLP} & \textbf{CBAS} & \textbf{WKB} & \textbf{PMD} \\
        \midrule
        SCP & 0.61\std{.04} & 0.55\std{.04} & 0.61\std{.03} & 0.49\std{.04} & 0.57\std{.03} & 0.71\std{.02} & 0.70\std{.03} & 0.72\std{.01} & 0.69\std{.02} & 0.75\std{.01} & 0.52\std{.06} & 0.61\std{.04} & 0.71\std{.01} \\
        NAPS & - & - & - & - & - & - & -& 0.82\std{.01} & 0.94\std{.01} & 0.96\std{.00} & 1.00\std{.00} & 1.00\std{.00} & 0.97\std{.00} \\
        DAPS & 0.52\std{.06} & 0.43\std{.05} & 0.56\std{.03} & 0.48\std{.04} & 0.50\std{.04} & 0.65\std{.03} & 0.56\std{.04} & 0.72\std{.02} & 0.70\std{.02} & 0.74\std{.01} & 0.52\std{.06} & 0.69\std{.03} & 0.70\std{.01} \\
        RLCP & 0.61\std{.04} & 0.53\std{.04} & 0.61\std{.03} & 0.49\std{.04} & 0.57\std{.03} & 0.71\std{.02} & 0.69\std{.03} & 0.72\std{.01} & 0.69\std{.02} & 0.75\std{.01} & 0.51\std{.06} & 0.60\std{.04} & 0.70\std{.01} \\
        CF-GNN & 0.61\std{.04} & 0.62\std{.04} & 0.67\std{.02} & 0.62\std{.04} & 0.63\std{.04} & 0.72\std{.02} & 0.74\std{.03} & 0.52\std{.08} & 0.64\std{.08} & 0.68\std{.07} & 0.61\std{.05} & 0.61\std{.07} & 0.63\std{.04} \\
        SNAPS & 0.67\std{.04} & 0.62\std{.03} & 0.64\std{.03} & 0.79\std{.04} & 0.58\std{.03} & 0.72\std{.03} & - & 0.73\std{.02} & 0.70\std{.02} & 0.75\std{.01} & 0.56\std{.06} & 0.68\std{.02} & 0.71\std{.01} \\
        RR-Mondrian & 0.64\std{.03} & 0.69\std{.03} & 0.71\std{.03} & 0.58\std{.04} & 0.71\std{.02} & 0.69\std{.02} & 0.67\std{.02} & 0.74\std{.02} & 0.73\std{.02} & 0.75\std{.01} & 0.59\std{.05} & 0.64\std{.06} & 0.79\std{.01} \\
        CalLCP & 0.74\std{.03} & 0.70\std{.04} & 0.80\std{.02} & 0.64\std{.03} & 0.75\std{.03} & 0.76\std{.02} & 0.79\std{.02} & OOT & 0.73\std{.01} & OOT & 0.62\std{.05} & 0.68\std{.03} & 0.79\std{.01} \\
        \midrule
        \ourmethod & 0.68\std{.03} & 0.71\std{.02} & 0.75\std{.01} & 0.68\std{.01} & 0.75\std{.01} & 0.77\std{.01} & 0.73\std{.03} & 0.80\std{.01} & 0.72\std{.01} & 0.81\std{.00} & 0.63\std{.02} & 0.70\std{.01} & 0.80\std{.01} \\
        \bottomrule
    \end{tabular}}
    \label{tab:wsc_coverage_all}
\end{table*}

\begin{table*}[t]
\scriptsize
\centering
\caption{WSC prediction length results. \ourmethod yields shorter relative prediction lengths on regression tasks than on classification tasks. OOT (out-of-time) shows results attained after an unreasonably long time.}
\vspace{-1mm}
\resizebox{\textwidth}{!}{
    \begin{tabular}{L{0.6in}C{0.3in}C{0.3in}C{0.3in}C{0.3in}C{0.3in}C{0.3in}C{0.3in}C{0.3in}C{0.3in}C{0.3in}C{0.3in}C{0.3in}C{0.3in}}
        \toprule
        \textbf{Method} & \textbf{ANH} & \textbf{CHI} & \textbf{EDU} & \textbf{ELC} & \textbf{INC} & \textbf{UNM} & \textbf{TWT} & \textbf{CMP} & \textbf{CRA} & \textbf{DBLP} & \textbf{CBAS} & \textbf{WKB} & \textbf{PMD} \\
        \midrule
        SCP & 1.66\std{.09} & 1.94\std{.07} & 2.02\std{.10} & 1.01\std{.03} & 1.94\std{.07} & 1.81\std{.04} & 2.10\std{.07} & 1.03\std{.04} & 0.91\std{.04} & 1.40\std{.06} & 1.99\std{.10} & 3.09\std{.34} & 1.22\std{.02} \\
        NAPS & - & - & - & - & - & - & - & 2.72\std{.23} & 5.18\std{.20} & 3.00\std{.10} & 3.78\std{.04} & 5.00\std{.00} & 2.12\std{.10} \\
        DAPS & 1.48\std{.08} & 1.75\std{.06} & 1.77\std{.09} & 1.01\std{.03} & 1.73\std{.06} & 1.58\std{.04} & 1.60\std{.05} & 1.41\std{.10} & 0.93\std{.04} & 1.37\std{.05} & 2.26\std{.08} & 4.07\std{.10} & 1.27\std{.02} \\
        RLCP & 1.66\std{.09} & 1.91\std{.07} & 2.02\std{.10} & 1.01\std{.03} & 1.94\std{.07} & 1.81\std{.04} & 2.10\std{.07} & 1.03\std{.04} & 0.90\std{.04} & 1.39\std{.06} & 1.99\std{.10} & 3.06\std{.33} & 1.22\std{.02} \\
        CF-GNN & 1.82\std{.22} & 1.67\std{.14} & 1.88\std{.12} & 0.85\std{.02} & 1.80\std{.07} & 1.61\std{.05} & 2.26\std{.07} & 3.68\std{.58} & 1.55\std{.30} & 2.01\std{.26} & 2.56\std{.29} & 3.78\std{.24} & 1.41\std{.21} \\
        SNAPS & 1.70\std{.09} & 2.10\std{.04} & 2.13\std{.03} & 1.23\std{.03} & 1.95\std{.04} & 1.81\std{.05} & - & 1.37\std{.12} & 0.91\std{.04} & 1.35\std{.04} & 2.78\std{.16} & 4.20\std{.05} & 1.19\std{.01} \\
        RR-Mondrian & 1.75\std{.23} & 1.78\std{.08} & 1.53\std{.24} & 0.51\std{.08} & 1.37\std{.20} & 1.41\std{.26} & 2.58\std{.58} & 3.27\std{.40} & 2.40\std{.34} & 1.96\std{.07} & 2.20\std{.25} & 2.65\std{.33} & 2.11\std{.05} \\
        CalLCP & 2.31\std{.26} & 2.33\std{.09} & 2.62\std{.14} & 1.10\std{.03} & 2.49\std{.08} & 2.17\std{.07} & 2.54\std{.14} & OOT & 1.39\std{.07} & OOT & 1.60\std{.22} & 4.14\std{.08} & 1.67\std{.03} \\
        \midrule
        \ourmethod & 1.66\std{.12} & 2.17\std{.14} & 2.44\std{.24} & 1.09\std{.06} & 2.52\std{.21} & 1.85\std{.05} & 2.10\std{.17} & 4.41\std{.06} & 2.00\std{.04} & 1.90\std{.08} & 2.75\std{.09} & 4.21\std{.06} & 1.93\std{.03} \\
        \bottomrule
    \end{tabular}}
    \label{tab:wsc_pred_length_all}
\end{table*}

\begin{table*}[t]
\scriptsize
\centering
\caption{Marginal prediction length. OOT (out-of-time) shows results attained after an unreasonably long time.}
\vspace{-1mm}
\resizebox{\textwidth}{!}{
    \begin{tabular}{L{0.6in}C{0.3in}C{0.3in}C{0.3in}C{0.3in}C{0.3in}C{0.3in}C{0.3in}C{0.3in}C{0.3in}C{0.3in}C{0.3in}C{0.3in}C{0.3in}}
        \toprule
        \textbf{Method} & \textbf{ANH} & \textbf{CHI} & \textbf{EDU} & \textbf{ELC} & \textbf{INC} & \textbf{UNM} & \textbf{TWT} & \textbf{CMP} & \textbf{CRA} & \textbf{DBLP} & \textbf{CBAS} & \textbf{WKB} & \textbf{PMD} \\
        \midrule
        SCP & 1.66\std{.09} & 1.94\std{.07} & 2.02\std{.10} & 1.01\std{.03} & 1.94\std{.07} & 1.81\std{.04} & 2.10\std{.07} & 1.08\std{.01} & 0.98\std{.01} & 1.16\std{.01} & 1.92\std{.03} & 4.26\std{.10} & 1.09\std{.00} \\
        NAPS & - & - & - & - & - & - & 5.73\std{1.18} & 2.19\std{.06} & 5.54\std{.06} & 3.26\std{.01} & 3.79\std{.03} & 5.00\std{.00} & 2.65\std{.01} \\
        DAPS & 1.48\std{.08} & 1.75\std{.06} & 1.77\std{.09} & 1.01\std{.03} & 1.73\std{.06} & 1.58\std{.04} & 1.60\std{.05} & 1.26\std{.02} & 0.99\std{.01} & 1.16\std{.01} & 2.05\std{.04} & 4.41\std{.06} & 1.12\std{.01} \\
        RLCP & 1.66\std{.09} & 1.91\std{.07} & 2.02\std{.10} & 1.01\std{.03} & 1.94\std{.07} & 1.81\std{.04} & 2.10\std{.07} & 1.08\std{.01} & 0.98\std{.01} & 1.16\std{.01} & 1.92\std{.03} & 4.24\std{.10} & 1.09\std{.00} \\
        CF-GNN & 1.82\std{.22} & 1.67\std{.14} & 1.88\std{.12} & 0.85\std{.02} & 1.80\std{.07} & 1.61\std{.05} & 2.26\std{.07} & 3.56\std{.63} & 1.55\std{.26} & 1.82\std{.25} & 2.40\std{.29} & 4.41\std{.15} & 1.30\std{.13} \\
        SNAPS & 1.70\std{.09} & 2.10\std{.04} & 2.13\std{.03} & 1.23\std{.03} & 1.95\std{.04} & 1.81\std{.05} & - & 1.22\std{.05} & 0.98\std{.01} & 1.16\std{.01} & 2.40\std{.03} & 4.32\std{.04} & 1.09\std{.00} \\
        RR-Mondrian & 2.12\std{.29} & 1.70\std{.07} & 2.18\std{.16} & 1.21\std{.07} & 2.05\std{.11} & 2.30\std{.19} & 4.61\std{.99} & 2.62\std{.13} & 2.36\std{.29} & 1.71\std{.02} & 2.73\std{.21} & 4.41\std{.11} & 1.43\std{.01} \\
        CalLCP & 2.24\std{.28} & 2.15\std{.06} & 2.24\std{.14} & 1.06\std{.02} & 2.12\std{.08} & 2.13\std{.06} & 2.54\std{.13} & OOT & 1.54\std{.03} & OOT & 2.71\std{.02} & 4.59\std{.03} & 1.15\std{.01} \\
        \midrule
        \ourmethod & 1.70\std{.11} & 1.92\std{.08} & 2.01\std{.10} & 1.03\std{.03} & 1.97\std{.06} & 1.84\std{.05} & 2.09\std{.15} & 4.33\std{.01} & 2.10\std{.02} & 1.81\std{.01} & 2.81\std{.04} & 4.45\std{.02} & 1.82\std{.01} \\
        \bottomrule
    \end{tabular}}
    \label{tab:marginal_pred_length_all}
\end{table*}

\begin{figure}[t]
    \centering
    \includegraphics[
        width=0.8\linewidth, 
        trim={0.2cm 0.2cm 0.2cm 0.0cm}, % trim={<left> <bottom> <right> <top>}
        clip                    % 'clip' is required for trim to work
    ]{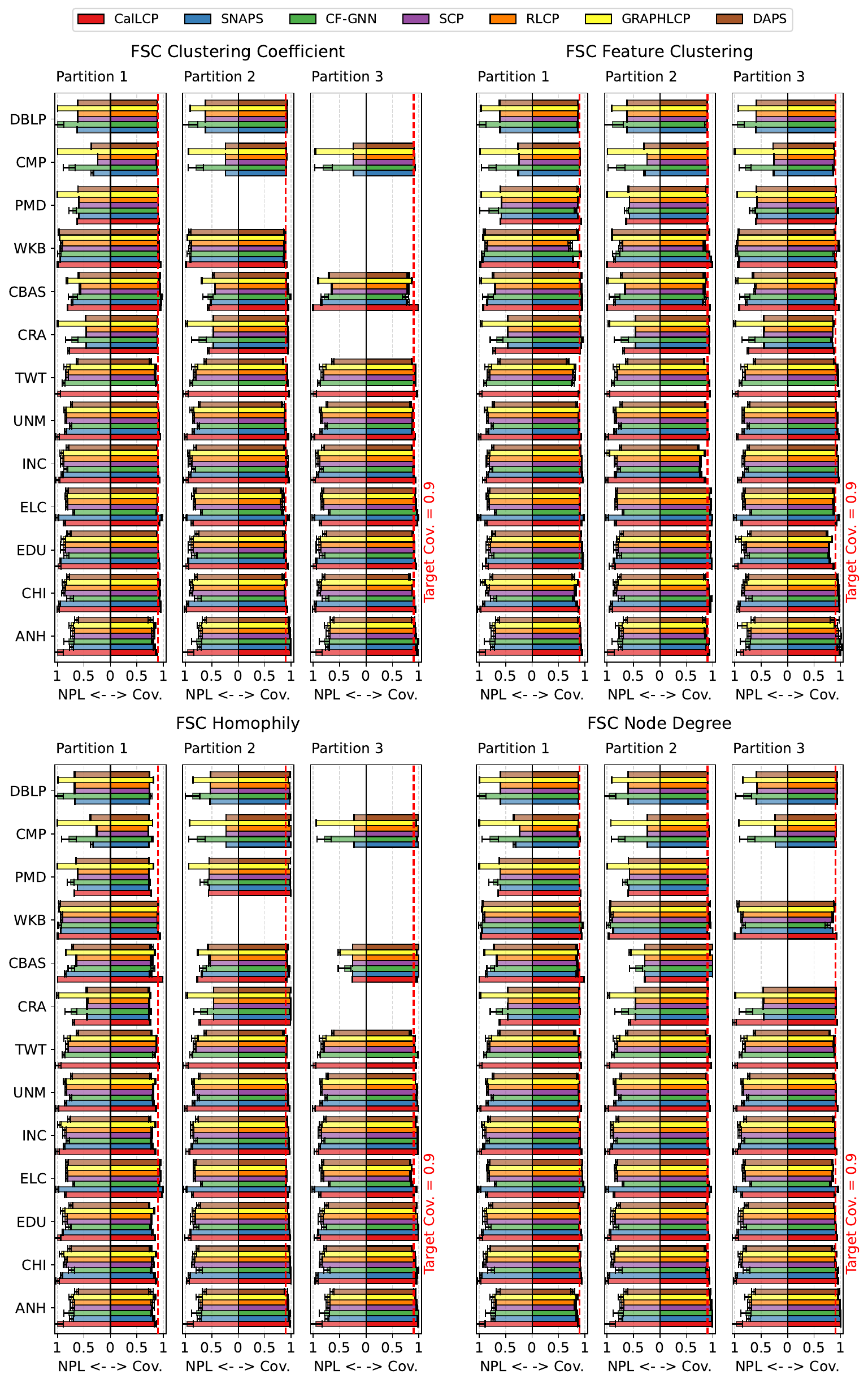}
    
    \caption{Results for group-based conditional coverage. It shows the minimum coverage across partitions for semantic and structural features where $\alpha=0.1$. \ourmethod demonstrates superior coverage compared to baselines and more specifically in homophily and feature clustering.}
    \label{fig:group_based_bar_all}
\end{figure}

\paragraph{Group-based conditional results} We provide additional conditional coverage results for all four types of group-based conditional coverage in Figure \ref{fig:group_based_bar_all}. 

\begin{table}[h]
\centering
\caption{Wall-Clock Runtime on Larger Datasets (seconds)}
\label{tab:runtime_large_datasets}
\resizebox{0.4\textwidth}{!}{\begin{tabular}{lccc}
\toprule
\textbf{Dataset} & \textbf{SCP} & \textbf{RLCP} & \textbf{GRAPHLCP} \\
\midrule
% CS   & 224.3 & 353.7 & 1066.1 \\
CMP  & 4.4   & 29.7  & 111.1  \\
DBLP & 7.5   & 28.3  & 508.8  \\
PMD  & 9.2   & 30.7  & 145.2  \\
% PHT  & 1.6   & 4.6   & 73.7   \\
% CSR  & 0.6   & 1.3   & -      \\
\bottomrule
\end{tabular}}
\end{table}

\paragraph{Marginal Prediction Length} Table \ref{tab:marginal_pred_length_all} shows that SCP generally yields short prediction sets, while CalLCP is often more conservative. \ourmethod produces shorter intervals than CalLCP across the regression datasets, alongside improved WSC over SCP and RLCP (Table \ref{tab:wsc_coverage_all}). Considering Table \ref{tab:marginal_coverage_all}, these comparisons should also account for CalLCP's frequent marginal overcoverage.

\subsection{Empirical runtime}\label{app:runtime}

We evaluate the wall-clock runtime of \ourmethod against SCP and RLCP on larger datasets. Results are in Table \ref{tab:runtime_large_datasets}. As expected, \ourmethod incurs noticeable overhead due to densification and PPR-based weighting, especially on DBLP. The fixed similarity threshold can limit added edges, but the additional cost remains a trade-off for improved localization. The results use the same hyperparameters as the main experiments. 

\subsection{Datasets}\label{app:datasets}
We show the detailed statistics of all datasets in Table \ref{tab:datasets}. We evaluate our proposed method across a diverse set of benchmark datasets across multiple domains and task types. For citation networks, we include Cora\_ML (CRA), DBLP, and PubMed (PMD), while the Amazon co-purchase domain is represented by Computers (CMP). We utilize a set of geographic and transportation datasets for regression tasks, including Education (EDU), Election (ELC), Income (INC), Unemployment (UNM), Anaheim (ANH), and Chicago (CHI). Finally, we incorporate the Twitch (TWT) social network and finally GOOD-WebKB (WKB) and GOOD-CBAS (CBAS).

\begin{table*}[h]
    \caption{Dataset Statistics}
    \small
    \centering

    \begin{tabular}{lccccc}
        \toprule
        \textbf{Dataset} & \textbf{Domain} & \textbf{\#Nodes} & \textbf{\#Edges} & \textbf{\#Features} & \textbf{\#Classes} \\
        \midrule
        Cora\_ML \cite{citationful2018}       & Citation       & 2,995   & 16,316  & 2,879  & 7  \\
        % CiteSeer \cite{citationful2018}      & Citation       & 4,230   & 10,674  & 602    & 6  \\
        DBLP \cite{citationful2018}           & Citation       & 17,716  & 105,734 & 1,639  & 4  \\
        PubMed \cite{citationful2018}         & Citation       & 19,717  & 88,648  & 500    & 3  \\
        Computers \cite{coauthor2019}      & Amazon         & 13,752  & 491,722 & 767    & 10 \\
        % Photo \cite{coauthor2019}         & Amazon         & 7,650   & 238,162 & 745    & 8  \\
        % CS \cite{coauthor2019}             & Coauthor       & 18,333  & 163,788 & 6,805  & 15 \\
        Education \cite{residualgnn2020}       & Geography      & 3,234   & 12,717  & 6      & Reg.  \\
        Election \cite{residualgnn2020}       & Geography      & 3,234   & 12,717  & 6      & Reg.  \\
        Income \cite{residualgnn2020}          & Geography      & 3,234   & 12,717  & 6      & Reg.  \\
        Unemployment \cite{residualgnn2020}    & Geography      & 3,234   & 12,717  & 6      & Reg.  \\
        Anaheim \cite{residualgnn2020}         & Transportation & 914     & 3,881   & 4      & Reg.  \\
        Chicago \cite{residualgnn2020}        & Transportation & 2,176   & 15,104  & 4      & Reg.  \\
        Twitch \cite{residualgnn2020}         & Social         & 1,912   & 31,299  & 3,170  & Reg.  \\
        GOOD-WebKB \cite{good2022}     & OOD Benchmark  & 617     & 1138       & 1703      & 5  \\
        GOOD-CBAS \cite{good2022}      & OOD Benchmark  & 700     & 3962       & 4      & 4 \\
        \bottomrule
    \end{tabular}
    \label{tab:datasets}
\end{table*}

\end{document}